\documentclass[journal]{IEEEtran}
\usepackage{amsmath,amsfonts}
\usepackage{algorithm, algorithmic}
\usepackage{array}
\usepackage{subfigure}
\usepackage{textcomp}
\usepackage{stfloats}
\usepackage{url}
\usepackage{multirow}
\usepackage{hyperref}
\usepackage{booktabs}
\usepackage{verbatim}
\usepackage{graphicx}
\usepackage{cite}
\usepackage{xcolor}
\begin{document}

\title{Constructing Enhanced Mutual Information for Online Class-Incremental Learning}

\author{\IEEEauthorblockN{
Huan Zhang, 
Fan Lyu,~\IEEEmembership{Member,~IEEE}, Shenghua Fan, Yujin Zheng, Dingwen Wang\IEEEauthorrefmark{2}}

\thanks{H. Zhang,  S. Fan, Y. Zheng and D. Wang, are with the School of Computer Science, Wuhan University, Wuhan, 430000, Hubei, China. Contacts: \texttt{\{cszhanghuan, fanshenghua, zhengyujin, wangdw\}@whu.edu.cn}}
\thanks{F. Lyu is with the New Laboratory of Pattern Recognition (NLPR), Institute of Automation, Chinese Academy of Sciences (CASIA), Beijing, 100190 China. Contacts: \texttt{fan.lyu@cripac.ia.ac.cn}}
\thanks{$\dagger$ Corresponding author.}
}
\markboth{Journal of \LaTeX\ Class Files,~Vol.~14, No.~8, August~2021}%
{Shell \MakeLowercase{et al.}: A Sample Article Using IEEEtran.cls for IEEE Journals}

\maketitle

\begin{abstract}

Online Class-Incremental continual Learning (OCIL) addresses the challenge of continuously learning from a single-channel data stream, adapting to new tasks while mitigating catastrophic forgetting.
Recently, Mutual Information (MI)-based methods have shown promising performance in OCIL.
However, existing MI-based methods treat various knowledge components in isolation, ignoring the knowledge confusion across tasks. 
This narrow focus on simple MI knowledge alignment may lead to old tasks being easily forgotten with the introduction of new tasks, risking the loss of common parts between past and present knowledge.
To address this, we analyze the MI relationships from the perspectives of diversity, representativeness, and separability, and propose an Enhanced Mutual Information (EMI) method based on knwoledge decoupling. 
EMI consists of Diversity Mutual Information (DMI), Representativeness Mutual Information (RMI) and Separability Mutual Information (SMI). 
DMI diversifies intra-class sample features by considering the similarity relationships among inter-class sample features to enable the network to learn more general knowledge. 
RMI summarizes representative features for each category and aligns sample features with these representative features, making the intra-class sample distribution more compact.
SMI establishes MI relationships for inter-class representative features, enhancing the stability of representative features while increasing the distinction between inter-class representative features, thus creating clear boundaries between class. 
Extensive experimental results on widely used benchmark datasets demonstrate the superior performance of EMI over state-of-the-art baseline methods.
\end{abstract}


\section{Introduction}
\begin{figure}
    \centering
    \includegraphics[width=1\linewidth]{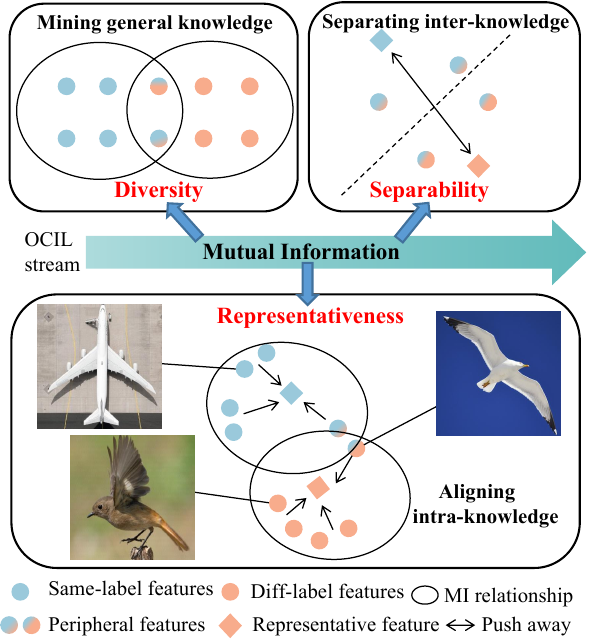}
    \caption{EMI considers the diversity, representativeness, and separability of samples. By enhancing the network's learning ability in OCIL from these three perspectives, EMI achieves better coupling of new and old knowledge.
 }
    \label{fig:intro}
\end{figure}

\IEEEPARstart{C}{lass}-incremental Learning (CIL) is a novel task that continually learns new classes from a series of tasks, where each task introduces new classes knowledge without associated task identifiers.
CIL is practical in real-world applications such as autonomous driving~\cite{autodrive}, remote sencing~\cite{tcsvt_remote} and robotics~\cite{robotic1}.
The goal of CIL is to accurately classify all categories from previous tasks without task-specific information while effectively managing Catastrophic Forgetting (CF) \cite{cf1,cf3, cf4}.
CF occurs when a model updates for new tasks and tends to overwrite previously acquired knowledge, leading to performance degradation. 
Recently, Online CIL (OCIL) task arises attention to meet the requirement that models have to adapt to constantly changing data streams with limited storage and computational resources. 
In OCIL, the model incrementally learns from a single-pass data stream. 

To alleviate CF, existing OCIL methods are generally divided into three categories. 
First, the regularization-based methods \cite{regular_1, tcsvt_regular2, tcsvt_regular1} enhance network stability by imposing additional constraints on parameters, thus reducing forgetting. Second, the architecture-based methods \cite{arch_1, arch_2, arch_3} dynamically adjust or modify the network structure to preserve previously acquired knowledge. 
Third, the replay-based methods \cite{chaudhry2019ER, aljundi2019mir, tcsvt_replay} maintain a memory buffer that stores samples from past tasks to retain old information. 
Although these methods have shown some effectiveness, in the context of online sequential learning, where each data point only be trained once from the single-passed data stream, the knowledge may be learned insufficiently.
The insufficient training leads to difficulties in effectively extracting knowledge from new tasks while knowledge from old tasks is even more prone to being forgotten. 

Recently, researchers have explored how Mutual Information (MI)~\cite{EstimatingMI, zhang2023mfsjmi, vinh2009information} can be used to address insufficient knowledge learning. 
MI is an information-theoretic metric to quantify the amount of information shared between two variables and excel at detecting both linear and non-linear dependencies.
Some existing methods have leveraged this property in combination with replay-based strategies to design MI-based OCIL approaches, such as OCM  \cite{guo2022ocm} and OnPro \cite{wei2023onpro}. 
OCM holds that cross-entropy loss is easy to learn biased features in an online mode, worsening catastrophic forgetting. 
To address this, OCM uses MI to build intra-class relationships, promoting comprehensive representations learning. 
OnPro \cite{wei2023onpro} further advances this approach by pushing away each class's online prototypes, thereby enhancing the decision boundaries of all seen classes.
{However, existing MI-based OCIL mehtods, such as OCM and OnPro, treat the knowledge within OCIL as isolation, ignoring the knowledge confusion in the online fashion.
Only building MI between old and novel tasks, which may lead to old tasks being easily overridden and forgotten with the introduction of new tasks and new tasks have poor plasticity.
}

\textit{This paper studies to solve the drawback of MI in OCIL, and proposes an Enhanced Mutual Information (EMI) framework based on knowledge decoupling. }
Specifically, EMI decouples knowledge to three kinds, say diversity, representativeness, and separability.
Knowledge with \textbf{Diversity} refers to the broad distribution and variation of samples from different tasks and classes within the feature space.
Knowledge with \textbf{Representativeness} focuses on identifying  the core characteristics of each class.
Knowledge with \textbf{Separability} refers to the ability of a model to distinguish and process data from different tasks when learning multiple tasks.
{The three kinds of knowledge are important for OCIL, because the data streams are often complex and characterized by intricate non-linear dependencies, which leads to confused decision and CF. }
For example, as illustrated in Fig. \ref{fig:intro}, among bird samples, birds flying in a blue sky may share many features with airplanes, such as white wings and similar flying shapes. This demonstrates the diversity of features intra- and inter-class. Although intra-class samples vary, they share more common features, such as bird feathers and tails, which strictly distinguish birds from airplanes. This highlights the representativeness intra-class and the separability inter-class.

Motivated by this, our EMI contains three key MI calculations, including Diversified Mutual Information (DMI), Representativeness Mutual Information (RMI), and Separability Mutual Information (SMI).
First, DMI enhances intra-class feature learning through inter-class samples, aiming to learn a richer set of features, thus improving the generalization capabilities.
Second, RMI leverages prototypes to summarize the representative features of each category. By maximizing the MI between samples and prototypes, RMI achieves a more cohesive intra-class sample distribution, thereby enhancing the representativeness for each OCIL task.
Third, SMI increases the inter-class distribution distance by constructing separability-enhanced MI for prototypes and their augmented views.
By emphasizing diversity, representativeness, and separability, our proposed EMI framework achieves a more uniform feature distribution across temporal data streams. 
This approach significantly boosts learning capabilities in OCIL, ensuring that models not only retain previously acquired knowledge but also effectively integrate new information.

The contributions of this paper are summarized as follows: 
 \begin{enumerate}
    \item To the best of our knowledge, we are the first to study decoupling knowledge with diversity, representativeness, and separability to advance the capabilities of MI learning in OCIL.
    \item We propose an Enhanced Mutual Information (EMI) framework based on DualNet's fast and slow learning theory. This framework uses fast and slow features to construct three types of MI to enhance OCIL: Diversified Mutual Information (DMI), Representativeness Mutual Information (RMI), and Separability Mutual Information (SMI).
    \item We conduct extensive experiments on three datasets, and the empirical results consistently demonstrate the superiority of our EMI over various state-of-the-art methods. Additionally, we investigate and analyze the benefits of each component through ablation studies, further confirming the effectiveness of our approach.
\end{enumerate}

\section{Related Work}
\subsection{Continual Learning}
Continual Learning focuses on enable models to learn continuously from a stream of data over time. Due to the restriction of network access to past data, catastrophic forgetting acts as a barrier to performance. To mitigating this issue, recent relevant methods can be categorized into three categories.
1) \textit{Regularization-based methods}, also known as prior-based methods, store previous knowledge learned from old data as prior information for the network. These methods \cite{regular_2,regular_3, regular_4, regular_5, ewc, lwf, regu_6} use historical knowledge to consolidate past knowledge by extending the loss function with an additional regularization term. For example, Chaudhry et al.~\cite{ewc} and its more advanced versions \cite{ewc_2, ewc3} using the Fisher information matrix to evaluate the importance of parameters.
2) \textit{Architecture-based methods}, also known as parameter-isolation methods, divide each task into a set of specific model parameters. This kind of methods \cite{arch_4,arch_5, arch_6} dynamically extend the model as the number of tasks increases or gradually freeze parts of the parameters to address the forgetting problem. For example, Kang et al.~\cite{kang2022forget} and Jin et al.~\cite{jin2022helpful} explicitly identifies the important neurons or parameters for each task and (almost) freezes them when training a new task. 
Ye et al.~\cite{ye2023self} dynamically expands the architecture through a self-assessment mechanism evaluating the diversity of knowledge among existing experts as expansion signals.
3) \textit{Replay-based methods}, which mitigate forgetting by maintaining  a fixed-size memory bank or synthesizing pseudo-data for pseudo-rehearsal ,also is named as rehearsal-based methods. This kind of methods \cite{ shim2021ASER, riemer2018mer, chaudhry2021using, ho2023prototype-guided, gao2023ddgr, gu2024summarizing, mai2021SCR} that replay old samples in the buffer are still the most effective for anti-forgetting at present.  Chaudhry et al.~\cite{chaudhry2021using} improves replay by adding an objective term to alleviate forgetting on the meta-learned anchor datapoints. Gao et al.~\cite{gao2023ddgr} adopts a diffusion model as the generator and calculates an instruction-operator through the classifier to instruct the generation of samples.

\subsection{Online Class-incremental Continual Learning}
Online continual learning focuses on a more practical setting where the  data stream is passed only once for training. Most online CL methods are based on replay. 
ER \cite{chaudhry2019ER} is one of the earliest replay-based method, which randomly retrieves samples from the memory bank and updates the bank using reservoir sampling. Follwing this, researchers use the memory retrieval strategy to select valuable samples from memory. ASER \cite{shim2021ASER}  leverages the samples stored in the memory bank to restrict parameter updates during training. MIR \cite{aljundi2019mir} selects memory samples whose losses are most significantly affected by the anticipated updates to the parameters. In the meantime, some approaches \cite{gu2024summarizing,prabhu2020gdumb} focus on saving more effective samples to the memory, belonging to the memory update strategy. SSD \cite{gu2024summarizing} refines real image data into more informative samples by capturing essential training characteristics. GDUMB \cite{prabhu2020gdumb} updates the memory buffer in a greedy manner, ensuring a balanced distribution across different classes. During inference, it employs a model trained from scratch exclusively using the balanced contents of the memory buffer. In addition to considering the update and retrieval of memory data, some work has also considered data relationships. iCaRL \cite{rebuffi2017icarl}  allows only a limited number of class data to be present during training, enabling the progressive addition of new classes.  SCR \cite{mai2021SCR} explicitly encourages samples from the same class to cluster tightly in embedding space while pushing those of different classes further apart during replay-based training. PCR \cite{lin2023pcr} replaces the contrastive samples of anchors with corresponding proxies in the contrastive-based way. 

Recently, some approaches have considered data relationships and task relationships from the perspective of MI. OCM \cite{guo2022ocm} constructs MI relationships inner-class data to obtain overall data features and also builds MI between past task data and current task data to mitigate forgetting. OnPro \cite{wei2023onpro} follows OCM and proposes online prototype equilibrium from the perspective of inter-class distance to learn representative features against shortcut learning. Unlike OCM and OnPro, we construct MI for decoupling knowledge with diversity, representiveness and separability.

\subsection{Mutual Information}
MI is a critical concept that measures the dependency between two variables in machine learning. Its ability to capture all types of relationships between variables, not just linear connections, makes it highly valuable for tasks such as feature selection \cite{zhang2023mfsjmi} , clustering analysis\cite{vinh2009information}, and in deepening the understanding and optimization of deep learning models\cite{cheng2006discriminative, tishby2015dlib}.  For example, Kinney et al.~\cite{kinney2014equitability}  discussed the concept of equitability and its relationship with MI, providing insights into the selection of features and variables in complex datasets. 
Poole et al.~\cite{poole2019variational} provided an analysis of MI's variational bounds within variational inference, enhancing understanding of optimization in machine learning. Zhang et al.~\cite{zhang2023mfsjmi} optimizes feature selection for multi-label data by introducing joint MI and interaction weight, improving the handling of high-order relevance between features and label sets in high-dimensional multi-label data.

Recently, as deep learning has rapidly evolved, the application of MI methods within this area has increasingly caught the attention of researchers. Lei et al.~\cite{lei2023mutual} transformed the hourly segmented measured data into the matrix form as the input of the deep learning model for training With the help of MI correlation analysis. Suh et al.~\cite{suh2023long} integrated contrastive learning with logit adjustment and reinterprets long-tailed recognition tasks as maximization of MI between latent features and ground-truth labels, proposing a new loss function that excels on long-tailed recognition benchmarks. 
Although MI is a powerful tool, it is hard to estimate. Boudiaf et al.~\cite{boudiaf2020unifying} designed an estimator for assessing MI based on the Donsker-Varadhan representation and the f-divergence representation. Then they used MI to bridge cross-entropy and pairwise loss and proposed a method for few-shot learning by maximizing the MI between queries and their label predictions. Oord et al.~\cite{oord2018representation} tried to solve the problem by
optimizing an InfoNCE-type lower bound of MI. 

Building on the aforementioned efforts, MI methods have begun to be developed in continual learning. Guo et al.~\cite{guo2022ocm} employs InfoNCE as a proxy to calculate MI between samples, using it as a loss function to mitigate forgetting in OCIL. Wei et al.~\cite{wei2023onpro} achieves intra-class compactness and learn instance-wise representations through MI maximization. Li et al.~\cite{li2023variational} maximizes the MI between the outputs of previously learned and current networks using a variational lower bound approach in an information-theoretic framework for knowledge distillation.

\begin{figure*}
  \centering
  \includegraphics[width=.9\textwidth]{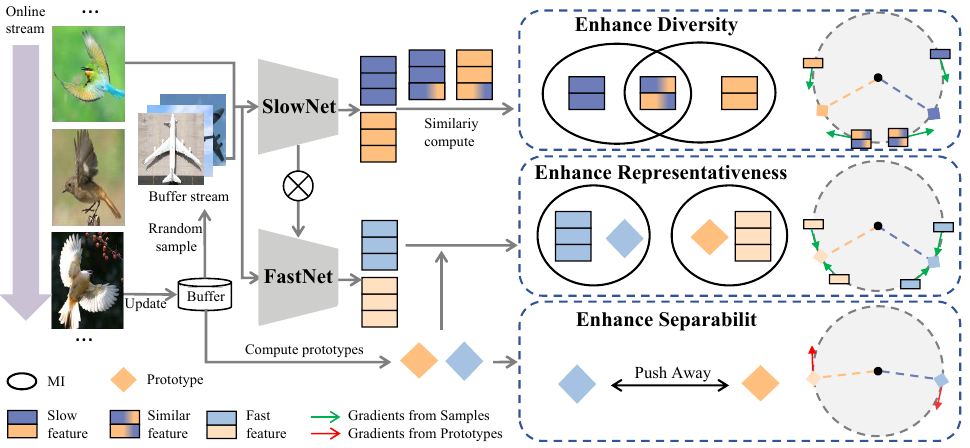}
  \caption{The proposed EMI framework leverages the complementary capabilities of DualNet. Initially, the incoming data stream and replay data are fed into the Slow Net to learn slow features. DMI utilizes these slow features to construct diversity-enhanced MI relationships. Subsequently, the slow features are transformed into fast features by the Fast Net. RMI and SMI leverage these fast features and prototypes to construct representativeness-enhanced and separability-enhanced MI, respectively. }
  \label{mainpng}
\end{figure*}

\section{Preliminaries}
\subsection{Problem Definition}
OCIL task is with the input of a continuous stream of data $\mathcal{D} = \{\mathcal{D}_1, \ldots, \mathcal{D}_T\}$, where $\mathcal{D}_t = { (x_{i}, y_{i}) }_{i=1}^{N_t}$ represents the dataset for a specific task $t$, with $T$ being the total number of tasks.
$N_t$ denotes the number of samples in dataset $\mathcal{D}_t$ and $y_{i}$ is the class label of sample $x_{i}$.
Each $y_{i}$ represents the class label of sample $x_{i}$, where $y_{i} \in \mathcal{C}_t$, with $\mathcal{C}_t$ being the class set for task $t$.
In OCIL, the class sets are disjoint, i.e., ${\mathcal{C}_{t_1}} \cap {\mathcal{C}_{t_2}} = \emptyset$ for any $i \neq j$.
Following replay strategy in other MI methods such as OCM, we also maintain a memory buffer $\mathcal{M}$ in our method following the replay strategy. 
At each time step of task $t$, the model receives a mini-batch data $\mathcal{X} \cup \mathcal{X}^b$ for training, where $\mathcal{X}$ and $\mathcal{X}^b$ are drawn from $\mathcal{D}_t$ and the memory buffer $\mathcal{M}$, respectively.  
In an online setting, each sample $(x_{i}, y_{i})$ is encountered just once within a non-stationary data stream, without any accompanying task information. The main objective of OCIL is to extract knowledge from the data stream and attain cross-task understanding.



\subsection{MI Estimation in OCIL}
\label{chap:MI Estimation}

In OCIL, each sample is trained only once, which may lead to insufficient training, which results in the model learning biased features that do not truly represent the actual distribution of the classes.
To tackle this issue, the Mutual Information (MI)~\cite{guo2022ocm} technique is proposed to establish intrinsic connections between tasks.
{MI can effectively mines complex and nonlinear relationships among variables, thus able to identifies the most significant features, reducing the dependence on bias. 
This property is important for the online setting, as it boosts the model's capacity to grasp the inherent structure of data, thus reducing biases from the rapid and irregular data flow.}
A commonly-used MI for online learning can be represents as follows.

For a mini-batch $\mathcal{X}$ in current stream, a simple MI can be computed between $\mathcal{X}$ and its augmentation view $\mathcal{X}'$ as a measure of the shared information. This can be written as:
\begin{equation}
\begin{aligned}
I(\mathcal{X};\mathcal{X}') &= \sum_{x\in\mathcal{X}, x'\in\mathcal{X}'}I(x;x') \\
&= \sum_{x\in\mathcal{X}, x'\in\mathcal{X}'} p(x|x') \log \frac{p(x|x')}{p(x)},
\label{mi1}
\end{aligned}
\end{equation}
where $I(x;x') $ indicating the dependency between two different samples in the batch $\mathcal{X}$.
Directly calculating MI is difficult on the raw data distribution $p(x)$. 
By utilizing the data processing inequality \cite{dataprocessinginequality1, dataprocessinginequality2}, the lower bound of MI is redefined on the feature space as:
\begin{equation}
\begin{aligned}
I(\mathcal{X};\mathcal{X}')  &\geq I(F(\mathcal{X});F(\mathcal{X}')) \\
&= \sum_{x\in\mathcal{X}, x'\in\mathcal{X}'} p(F(x)|F(x')) \log \frac{p(F(x)|F(x'))}{p(F(x))},
\label{mi2}
\end{aligned}
\end{equation}
where $F(\cdot)$ is a feature extractor.

However, the calculation of Eq.~\eqref{mi2} remains infeasible without access to $p(F(x))$. 
To address this issue, Oord \cite{oord2018representation} introduced InfoNCE as a proxy to maximize MI between $F(\mathcal{X})$ and its augment view $F(\mathcal{X}')$, which can be expressed as:
\begin{align}
I(F(\mathcal{X}); F(\mathcal{X}')) \geq \log N + \text{InfoNCE}(\mathcal{X},\mathcal{X}'), 
\label{eq:I(F(X), F(X'))}
\end{align}
where $N$ is the size of $\mathcal{X}$ and
\begin{align}
 \text{InfoNCE}(\mathcal{X},\mathcal{X}') = \sum_{i=1}^{N} \log  \frac{ <x_i, x_i'>}{\frac{1}{N} \sum_{j=1}^{N} <x_i, x_j'>}.
 \label{infonce}
\end{align}
$<x_i, x_j'> = e^{\frac{F(x_i)^T F(x_j')}{\tau}}$ is the similarity of two samples and $\tau$ is temperature. InfoNCE exhibits stronger adaptability to high-dimensional data and significantly improves computational efficiency by avoiding direct probability density estimation. 


Based on MI, some recent OCIL studies design MI relationships to improve the stability and plasticity in online sequential learning, such as OCM \cite{guo2022ocm}, InfoCL \cite{song2023infocl} and OnPro \cite{wei2023onpro}.
These methods build MI including between inputs and features, between features with labels and between past and present. 
{
However, these MI methods ignore that the knowledge within OCIL data stream is confused, directly leveraging  MI may amplify this confusion.
As illustrated in Fig. \ref{fig:intro}, when the network learns the features of the current task involving airplanes, some features of bird samples from previous tasks share similarities with airplane features. 
Ignoring these similarities and directly applying MI or other contrastive learning methods for forced partitioning can lead to the forgetting of old tasks and confusion in learning new tasks during insufficient online training.
}
To address this, we propose to decouple the knowledge with OCIL with MI, establishing finer-grained associations of knowledge expressed through three core MI calculations: diversity, representativeness, and separability.
{By constructing the enhanced MI relationships based on these three properties, we aim to achieve the coupling of old and new knowledge in OCIL, which allows for the retention of more previously learned knowledge while also enabling a reasonable distinction between old and new knowledge.
}

\section{Method}
\subsection{Overview}

This paper aims to enhance MI relationships in OCIL by knowledge decoupling into diversity, representativeness, and separability.
Specifically, we propose an Enhanced Mutual Information (EMI) method, which consists of three core components, including Diversified Mutual Information (DMI), Representative Mutual Information (RMI), and Separable Mutual Information (SMI).

To better decouple knowledge in an online fashion, we design EMI based on the DualNet structure~\cite{duannet}, which contains a Slow Net and a Fast Net with different architecture.
DualNet leverages a Slow Net to learn general cross-category shared slow features and a Fast Net to learn class-specific fast features.
The Slow Net has a ResNet with four residual blocks, while the Fast Net consists of four convolutional layers. Let $\{r^s_{i}\}_{i=1}^4$ represent the feature maps derived from the four residual blocks of the Slow Net for an input $x$, where $i$ denotes the output from each corresponding respective block. 
The DualNet framework utilizes the Fast Net to transform these slow features $\{r^s_{i}\}_{i=1}^4$ into fast features $\{r^f_{i}\}_{i=1}^4$ for classification purposes. Each convolutional layer in the Fast Net is directly aligned with a corresponding residual block in the Slow Net, ensuring a precise alignment in the feature transformation process. The fast features are obtained as follows:
\begin{align}
r^f_{i} &= f'_{i}(r^s_{i}) \odot r^s_{i},\quad i=1,2,3,4,
\label{eq:t(f)}
\end{align}
where $\odot$ denotes the element-wise multiplication, $f'_{i}$ denotes the $i$-th CNN layer of the Fast Net. $r^f_{i}$ has the same dimension as the corresponding $r^s_{i}$.
Specifically, for a sample $x$, the Slow Net $f$ extracts the slow feature $f(x)$, which the Fast Net $h$ then transforms into fast feature $h(f(x))$. 
The classifier $\phi$ uses this fast feature to predict the class label, $\phi(h(f(x)))$.
Both the slow and Fast Net incorporate projection heads, $g$ and $g'$, respectively. 
These heads project the features into a distinct space to establish MI relationships, represented as $F(x_i) = g(f(x_i))$ and $F'(x_i) = g'(h(f(x_i)))$.
In the following, we will introduce how to construct three kinds of MI to improve the OCIL task.

For the slow features from the Slow Net (denote as $f$), DMI incorporates cross-category similar features into the learning of intro-class features and encourages the network to learn more general intro-class features, sharing features across categories through diverse intro-class feature learning.
For the fast features from the Fast Net (denote as $h$), we develop Representative Mutual Information (RMI) and Separable Mutual Information (SMI). RMI constructs MI relationships for representative features within each class, aligning the features of intro-class samples with the representative features of the class. 
SMI promotes the separation between classes by enhancing the distinctions of representative features among classes, resulting in a clearer boundary between class distributions.
In the following, we will introduce DMI, RMI and SMI in detail.

\subsection{EMI for Diversified Knowledge Decoupling}
\label{chap:Enhancing Diversity in MI Construction}

{Diversified Knowledge refers to a broader and richer set of feature information, enabling the model to capture the underlying similarities between different tasks and categories~\cite{diversity}. 
The Slow Net is a crucial component of the DualNet framework, which is designed to learn generic, task-agnostic feature representations. In our study, DMI leverages the Slow Net to achieve knowledge decoupling for diversified knowledge.}

To achieve diversified knowledge within OCIL, we assume that samples from different tasks and classes exhibit latent similarities, and the diversified MI can be found in inter-class similar features.
Thus, we propose to harness the latent similarities among inter-class samples to enrich the Diversified Mutual Information (DMI). 
Given the slow features from DualNet, DMI construct diversified intra-class MI $I_\text{d}(\mathcal{X};\mathcal{S})$ to enrich intra-class diverse knowledge, and $I_\text{d}(\mathcal{X};\mathcal{S})$ can be constructed as follows:
\begin{equation}
I_\text{d}(\mathcal{X}; \mathcal{S})  \geq I_\text{d}(F(\mathcal{X}); \mathcal{S}) 
= H(\mathcal{S}) - H((F(\mathcal{X})|\mathcal{S}),
\label{eq:mixsb}
\end{equation}
where $\mathcal{S} = \{\mathcal{S}_0, \mathcal{S}_1, ..., \mathcal{S}_N\}$ represents the diversified set for a mini batch $\mathcal{X}$. 
For each $\mathcal{S}_i\in\mathcal{S}$ is the diversified set for sample $x_i\in\mathcal{X}$, which can be calculated as:
\begin{align}
\mathcal{S}_i = \{ j \in \{ 1,2,...,N \}|i \neq j, s_{ij} > \mu \cup y_{x_j} = y_{x_i} \},
\label{eq:compute_similarity}
\end{align}
where $s_{ij}= e^{\frac{F(x_i)^T F(x_j)}{\tau}}$ indicates the pairwise similarity of $F(x_i)$ and $F(x_j)$. $\mathcal{S}_i$ includes all samples of the intra-class as $x_i$ and inter-class similar samples in $\mathcal{X}$. 
Then, given the diversified set $\mathcal{S}$, we can easily compute DMI $I_\text{d}(\mathcal{X};\mathcal{S})$. 
However, it is challenging to directly optimize Eq .\eqref{eq:mixsb} because it is difficult to evaluate $H(\mathcal{X})$ and $H(\mathcal{X}|\mathcal{S})$.
For effective optimization, we also transform the Eq.~\eqref{eq:mixsb} into the InfoNCE~\cite{guo2022ocm} lower bound of $I(F(\mathcal{X}); \mathcal{S})$ for better optimization:
\begin{align} 
\ell(F(\mathcal{X}, \mathcal{S}) &= -\frac{1}{N} \sum_{i=1}^{N} \frac{\log A_i}{\sum_{l=1}^{N} \mathbb{I}(l \in \mathcal{S}_i)},
\label{eq:l(x,s)}
\end{align}
where
\begin{align} 
A_i &= \frac{\sum_{k \in \mathcal{S}_i}s_{ik} \cdot s_{ik'}  \cdot s_{i'k}}{\left(\sum_{j=1}^{N}s_{ij} +s_{ij'} +s_{i'j} \right)^3},
\end{align}
here $s_{ij'}$ denotes pairwise similarity between $F(x_i)$ and $F(x_j')$. 
In Eq. \eqref{eq:mixsb}, to maximize $I(\mathcal{X}; \mathcal{S})$, the entropy $H(\mathcal{S})$ should be maximized (the first term), and the label should be easily identified from the feature representation (the second term, which is naturally minimized by Eq. \eqref{eq:l(x,s)}. Maximizing $H(\mathcal{S})$ implies that the variable $\mathcal{S}$ should follow a uniform distribution. 
To avoid class imbalance caused by directly combining $\mathcal{X}$ and $\mathcal{X}^b$ (where classes in $\mathcal{X}^b$ may be overrepresented), we train $\mathcal{X}$ and $\mathcal{X}^b$ separately. The classes in $\mathcal{X}$ or $\mathcal{X}^b$ are approximately balanced when considered independently, ensuring a more stable training process and preventing class imbalance issues.
In summary, the DMI loss can be denoted as follows:
\begin{equation}
\begin{aligned}
&\mathcal{L}_\text{dmi}= \ell(F(\mathcal{X}), \mathcal{S}) + \ell(F(\mathcal{X}^b), \mathcal{S}^b)\\
&=-\frac{1}{N} \sum_{i=1}^{N} \frac{ \log A_i}{\sum_{l=1}^{N} \mathbb{I}{(l \in \mathcal{S}_i)}}-\frac{1}{M} \sum_{i=1}^{M} \frac{ \log A_i^b}{\sum_{l=1}^{M} \mathbb{I}{(l \in \mathcal{S}_i^b)}},
\label{eq:dmi}
\end{aligned}
\end{equation}
where
\begin{align} 
A_i^b &= \frac{\sum_{k \in \mathcal{S}_i}s_{ik}^b \cdot s_{ik'}^b  \cdot s_{i'k}^b}{\left(\sum_{j=1}^{N}s_{ij}^b +s_{ij'}^b +s_{i'j}^b \right)^3}.
\end{align}
The similarity set $\mathcal{S}^b$ is also computed with Eq. \eqref{eq:compute_similarity}.

{By the proposed DMI, we enable the Slow Net to decouple diversified knowledge and enhance the diversity in OCIL, . 
As illustrated in the top right part of Fig. \ref{mainpng}, DMI supplements the MI relationships with inter-class samples that are highly similar to intra-class samples, which enriches the diversity of intra-class knowledge, facilitating the learning of more marginal and variant features as well as shared knowledge of inter-class.
From a optimization perspective, sharing gradients of similar samples of inter-class allows the features of these samples to achieve more reasonable positions on the hypersphere. 
This results in a more uniform distribution of sample features at the feature level. 
Moreover, the diversified knowledge learned by the Slow Net is integrated into the Fast Net through the slow-fast transformation, providing generalized and robust initialization information for subsequent RMI and SMI to optimize fast features.}

\subsection{EMI for Representativeness and Separability Knowledge}

The Fast Net of DualNet integrates fast features to learn class-specific knowledge with stronger generalization capabilities. 
In this study, we utilize the Fast Net to decouple representativeness knowledge and separability knowledge. The former ensures the consistency of key intra-class knowledge, while the latter ensures the differentiation of key inter-class knowledge.
On top of this, we propose RMI and SMI for representativeness and separability knowledge.
For RMI, we align the fast features of all intra-class samples with the representative features of the class, yielding RMI $I_\text{d}(\mathcal{X};\mathcal{R})$, where $\mathcal{R}$ represents the class's representative features. 
For SMI, we emphasize the differences between inter-class representative features, and construct MI between the representative features and their augmented views, which denotes as $I_\text{s}(\mathcal{R}; \mathcal{R}')$. 

\subsubsection{Enhancing Representativeness in MI}
Representativeness refers to features that capture and encapsulate the core characteristics of a class~\cite{representativeness}. Constructing Enhanced RMI $I_\text{r}(\mathcal{X}; \mathcal{R})$ can align the fast features of intra-class samples with representative features within the feature space, thereby enhancing the consistency of intra-class samples. This alignment makes the intra-class features more compact, facilitating improved classification accuracy and robustness. RMI can be constructed as:
\begin{equation}
I_\text{r}(\mathcal{X}; \mathcal{R})  \geq I_\text{r}(F(\mathcal{X}); \mathcal{S}) 
= \text{InfoNCE}(\mathcal{X}, \mathcal{R}).
\end{equation}
However, directly acquiring feature $\mathcal{R}$ is challenging due to the difficulty in discerning representative features from complex and high-dimensional feature vectors. 

In this study, we employ prototypes $\mathcal{P}$ to construct RMI. 
To compute the prototype in OCIL, 
we randomly sample $N_\text{p}$ samples from the memory buffer for each seen class, with the sampled set denoted as $\mathcal{X}^p$. The prototype for each class is defined as the mean of the fast features of these samples, represented as follows:
\begin{align}
p_i = -\frac{1}{N_{p}} \sum_{j} F'(x_{j}) \cdot \mathbb{I}\{y_{p_i}= y_{x_j}\},
\label{eq:prototype}
\end{align}
where $x_j \in \mathcal{X}^p$, \( \mathbb{I} \) is the indicator function, and $p_i$ is the prototype of class $i$. 
In OCIL, we need to maintain a set of $K$ prototypes $\mathcal{P}=\{p_i\}^K_{i=1}$ and its augmented view $\mathcal{P}'=\{p_i'\}^K_{i=1}$ through augment view of $\mathcal{X}^p$, where $K$ represents the number of classes in $\mathcal{X}$. 
What's more, given that fast features integrate the more general representations of slow features, we aim to align the fast features with the representative features of their respective classes as closely as possible. This alignment ensures that the intra-class feature distribution becomes more compact.
To this end, we construct MI relationships between the prototypes and the fast features, so we rewrite $I_\text{r}(F(\mathcal{X});\mathcal{R})$ as $I_\text{r}(F'(\mathcal{X});\mathcal{P})$. Following Eq. \eqref{eq:l(x,s)}, we maximize $I_\text{r}(F'(\mathcal{X});\mathcal{P})$ through maximize its lower bound:
\begin{equation}
\begin{aligned}
\ell(F'(\mathcal{X}), \mathcal{P})= -\frac{1}{K} \sum_{i=1}^{K} \frac{\log B_i}{\sum_{l=1}^{N} \mathbb{I}(y_{x_l} = y_{p_i})}
,
\end{aligned}
\end{equation}
where
\begin{equation}
B_i =\\
\frac{\sum_{y_{x_k} = y_{p_i}}\langle p_i,F_t(x_k) \rangle \langle p_i,F_t(x_k') \rangle \langle p_i',F_t(x_k) \rangle}{\left(\sum_{j=1}^{N}\langle p_i,F_t(x_j) \rangle + \langle p_i,F_t(x_j') \rangle + \langle p_i',F_t(x_j) \rangle\right)^3}.
\end{equation}
We also derive the prototype for $\mathcal{X}^b$, denoted as $\mathcal{P} = \{p_i^b\}^{K^b}_{i=1}$, where $K^b$ represents the number of classes in $\mathcal{X}^b$. For the same reason in Eq. \eqref{eq:l(x,s)}, we train $\mathcal{X}$ and $\mathcal{X}^b$ separately and we can get our RMI loss:
\begin{equation}
\begin{aligned}
&\mathcal{L}_\text{rmi} = \ell(\mathcal{X}, \mathcal{P}) + \ell(\mathcal{X}^b, \mathcal{P}^b)\\
&= -\frac{1}{K} \sum_{i=1}^{K} \frac{\log B_i}{\sum_{l=1}^{N} \mathbb{I}(y_{x_l} = y_{p_i})} -\frac{1}{K^b} \sum_{i=1}^{K^b} \frac{\log B_i^b}{\sum_{l=1}^{M} \mathbb{I}(y_{x_l} = y_{p_i})},
\label{eq:rmi}
\end{aligned}
\end{equation}
where
\begin{equation}
B_i^b =\\
\frac{\sum_{y_{x_k}^b = y_{p_i}^b}\langle p_i^b,F_t(x_k^b) \rangle \langle p_i^b,F_t(x_k'^b) \rangle \langle p_i'^b,F_t(x_k^b) \rangle}{\left(\sum_{j=1}^{M}\langle p_i^b,F_t(x_j^b) \rangle + \langle p_i^b,F_t(x_j'^b) \rangle + \langle p_i'^b,F_t(x_j^b) \rangle\right)^3}.
\end{equation}

RMI enhances class-specific characteristics and improves the overall discriminative ability of the model. As illustrated in the right part of Fig. \ref{mainpng}, 
{We utilize the slow-to-fast transformation to integrate diversified knowledge into fast features, ensuring more robust fast features. RMI constructs MI between fast features and their corresponding key expressions, enhancing the consistency of intra-class feature expressions. This enables the network to learn pattern-separated representations that are highly consistent and possess stronger generalization capabilities for specific classes.} 
From a gradient perspective, RMI guides fast features to converge towards key expressions, making broader features more compact and consistent within classes. This process also establishes clearer boundaries between different classes, enhancing the model's discriminative ability.

\subsubsection{Enhancing Separability in MI Construction}
Separability refers to the ability of a model to distinguish and process data from different tasks when learning multiple tasks~\cite{separability}. Enhancing separability in the construction of MI can stabilize the expression of representative features while also increasing the differences between inter-class representative features. Therefore, we have constructed SMI $I_\text{s}(\mathcal{R};\mathcal{R}')$ for the representative features and their augmented views. This can be expressed as:
\begin{align}
I_\text{s}(\mathcal{R};\mathcal{R}') \geq \text{InfoNCE}(\mathcal{R},\mathcal{R}').
\end{align}
Similar to RMI, we use $\mathcal{P}$ to substitute $\mathcal{R}$, and we can formulate our optimization objective as follows\cite{wei2023onpro}: 
\begin{align}
\ell(\mathcal{P}, \mathcal{P}')=-
\frac{1}{\mathcal{P}} \sum_{i=1}^{\mathcal{P}} \log  \frac{ <p_i, p_i'>}{\frac{1}{K} \sum_{j=1}^{K} <p_i, p_j'>},
\label{mi_pp}
\end{align}
we train $\mathcal{X}$ and $\mathcal{X}^b$ sparately and the SMI loss is:
\begin{align}
\mathcal{L}_\text{smi} = \ell(\mathcal{P}, \mathcal{P}') + \ell(\mathcal{P}^b, \mathcal{P}'^b).
\label{eq:smi}
\end{align}
As illustrated in the lower right part of Fig. \ref{mainpng}, SMI pushes away prototypes of different classes, making the representative features of each class more distinct. It is important to emphasize that in the EMI framework, SMI supplements RMI. This ensures that the representative features with which intra-class samples align are more distinct, making it easier to differentiate between samples of different classes.

\begin{algorithm}[t]
\caption{\textbf{Enhanced Mutual Information}}
\label{alg:emi}
\begin{algorithmic}[1]
\REQUIRE Initialized Fast Net $f$ and $g$, Initializared Slow Net $h$ and $g'$, classifier $\phi$, memory $\mathcal{M}$, data stream $D_{1:T}$
\FOR{$t \leftarrow 1$ to $T$}
    \FOR{$\mathcal{X},\mathcal{Y}$ in $D_t$}

        \STATE Randomly sample replay data $\mathcal{X}^b,\mathcal{Y}^b$ from $\mathcal{M}$  \hfill
        
        \textcolor{blue}{// \textit{DualNet ouputs slow and fast feature:}}
        \STATE $F(\mathcal{X}) \leftarrow g(f(\mathcal{X})), F(\mathcal{X}^b) \leftarrow g(f(\mathcal{X}^b))$
        \STATE $F'(\mathcal{X}) \leftarrow g'(h(f(\mathcal{X}))),F'(\mathcal{X}^b) \leftarrow g'(h(f(\mathcal{X}^b)))$ \hfill
        
        \textcolor{blue}{// \textit{Diversity MI construction:}}
        \STATE $S,S^b \leftarrow$ \text{Compute diversified set as Eq. \eqref{eq:compute_similarity}} \\
        \STATE $\mathcal{L}_\text{dmi} \leftarrow \text{DMI$(F(\mathcal{X}),F(\mathcal{X}^b),S,S^b)$ as Eq. \eqref{eq:dmi}}$ \hfill

        \textcolor{blue}{// \textit{Representativeness MI construction:}}
        \STATE $\mathcal{P}, \mathcal{P}^b \leftarrow$ Compute prototype as Eq. \eqref{eq:prototype}
        \STATE $\mathcal{L}_\text{rmi} \leftarrow $RMI$(F'(\mathcal{X}),F'(\mathcal{X}^b),\mathcal{P},\mathcal{P}^b)$ as Eq .\eqref{eq:rmi} \hfill
        
        \textcolor{blue}{// \textit{Separability MI construction:}}
        \STATE $\mathcal{L}_\text{smi} \leftarrow \text{SMI$(\mathcal{P},\mathcal{P}^b)$ as Eq .\eqref{eq:smi}}$ \
        \STATE $\mathcal{L}_\text{ce} \leftarrow \text{CE}(\mathcal{Y}^b, \phi(h(f(\mathcal{X}^b))))$ \\
        \STATE $\mathcal{L}_\text{ocm} \leftarrow \text{Use fast feature as Eq. \eqref{eq:ocm}}$ \\
        \STATE Update $f, g, h, g', \phi$ by $\mathcal{L}_\text{dmi}, \mathcal{L}_\text{rmi}, \mathcal{L}_\text{smi}, \mathcal{L}_\text{ce}, \mathcal{L}_\text{ocm}$ \\
        \STATE $\mathcal{M} \leftarrow \text{MemoryUpdate}(\mathcal{M}, \mathcal{X}, \mathcal{Y})$
    \ENDFOR
\ENDFOR
\STATE \textbf{return} {$f, h, g, g', \phi$}
\end{algorithmic}
\end{algorithm}

\subsection{Put It All Together}
The overall training prosess of EMI is shown in the Algorithm .\ref{alg:emi}. EMI utilizes slow features and fast features generated by the DualNet to construct three different types of enhanced MI relationships for supervising network learning.

DMI utilizes slow features to learn richer features, thus the optimization goal of the Slow Net is:
\begin{equation}
\mathcal{L}_\text{slow} = \mathcal{L}_\text{dmi}.
\end{equation}

RMI and SMI utilize fast features to learn class-specific features. By enhancing representativeness and separability in MI construction, they effectively ensure a more uniform data distribution, thereby improving the model's ability to discriminate between different classes. However, the model does not yet account for intra-class MI relationships and current-past MI relationships. To address this, we employ additional MI relationships as outlined in OCM \cite{guo2022ocm} to these representations:
\begin{equation}
\begin{aligned}
\mathcal{L}_\text{ocm} &= \max \left\{ I(\mathcal{X} ; F(\mathcal{X})) + I(\mathcal{X}^{\text{b}} ; F(\mathcal{X}^{\text{b}})) \right. \\
& \quad + \left. I(F^{-1}(\mathcal{X}^{\text{b}}) ; F(\mathcal{X}^{\text{b}})) \right\},
\label{eq:ocm}
\end{aligned}
\end{equation}
where $F^{-1}$ means the past model of last task. To this end, the optimization goal of the Fast Net is:
\begin{equation}
\mathcal{L}_\text{fast}=\mathcal{L}_\text{rmi} + \mathcal{L}_\text{smi} +\mathcal{L}_\text{ce} + \mathcal{L}_\text{ocm}, 
\end{equation}
where $\mathcal{L}_\text{ce} = \text{CE}(y_i^b, \phi(h(f(x_i^b))))$ is the cross-entropy loss. To this end, we can obtain our final optimization objective:
\begin{equation}
\mathcal{L}=\mathcal{L}_\text{slow} + \mathcal{L}_\text{fast}.
\end{equation}
What's more, following other replay-based methods~\cite{chaudhry2019ER, lin2023pcr, mai2021SCR}, we update the memory bank in each time step by uniformly randomly selecting samples from $\mathcal{X}$ to push into $\mathcal{M}$ and, if $\mathcal{M}$ is full, pulling an equal number of samples out of $\mathcal{M}$.

\section{Experiment}
\subsection{Experiment Setup}

\begin{table*}[t]
\centering
\caption{Average Accuracy (higher is better) on three benckmark datasets (CIFAR-10 with 5 tasks, CIFAR-100 witch 10 tasks and Tiny-ImageNet with 100 tasks) with different memory sizes. All results are the average and standard deviation of 10 runs.}
\label{tab:ac}
\begin{tabular}{l|ccc|ccc|ccc}
\toprule
\multirow{2}{*}{\textbf{Method}} & \multicolumn{3}{c|}{\textbf{CIFAR-10} (5 tasks)} & \multicolumn{3}{c|}{\textbf{CIFAR-100} (10 tasks)} & \multicolumn{3}{c}{\textbf{Tiny-ImageNet} (100 tasks)} \\
                        & \(M=0.2k\) & \(M=0.5k\) & \(M=1k\) & \(M=0.5k\) & \(M=1k\) & \(M=2k\) & \(M=1k\) & \(M=2k\) & \(M=4k\) \\ \midrule
fine-tune &17.4\,\(\pm\)1.0 & 17.4\,\(\pm\)1.0 & 17.4\,\(\pm\)1.0 & 5.8\,\(\pm\)0.3 & 5.8\,\(\pm\)0.3 & 5.8\,\(\pm\)0.3 & 0.8\,\(\pm\)0.1 & 0.8\,\(\pm\)0.1& 0.8\,\(\pm\)0.1 \\
offline &81.5\,\(\pm\)0.3 & 81.5\,\(\pm\)0.3 & 81.5\,\(\pm\)0.3 & 50.2\,\(\pm\)0.3 & 50.2\,\(\pm\)0.3 & 50.2\,\(\pm\)0.3 & 25.9\,\(\pm\)0.4 & 25.9\,\(\pm\)0.4 & 25.9\,\(\pm\)0.4\\
\midrule
AGEM~\cite{chaudhry2018agem}                    &17.5\,\(\pm\)0.6 & 17.6\,\(\pm\)1.0 & 17.7\,\(\pm\)1.0 & 5.9\,\(\pm\)0.5 & 5.5\,\(\pm\)0.2 & 5.8\,\(\pm\)0.8 & 0.8\,\(\pm\)0.1 & 0.7\,\(\pm\)0.1 & 0.8\,\(\pm\)0.2\\

ER~\cite{shim2021ASER} & 20.9\,\(\pm\)0.5 & 25.0\,\(\pm\)0.5 & 31.3\,\(\pm\)1.5 & 9.3\,\(\pm\)0.2 & 10.9\,\(\pm\)0.2 & 15.4\,\(\pm\)0.1 & 1.2\,\(\pm\)0.1 & 1.4\,\(\pm\)0.2 & 2.7\,\(\pm\)0.2 \\
MIR~\cite{aljundi2019mir} & 27.6\,\(\pm\)0.6 & 33.5\,\(\pm\)0.4 & 44.7\,\(\pm\)0.8 & 9.7\,\(\pm\)0.1 & 12.1\,\(\pm\)0.1 & 15.7\,\(\pm\)0.2 & 1.3 \,\(\pm\)0.1 & 1.8\,\(\pm\)0.3 & 3.6\,\(\pm\)0.2 \\

GSS~\cite{aljundi2019gss} & 22.1\,\(\pm\)0.5 & 30.3\,\(\pm\)0.8 & 37.8\,\(\pm\)0.9 & 9.4\,\(\pm\)0.2 & 11.5\,\(\pm\)0.2 & 12.2\,\(\pm\)0.2 &  1.1\,\(\pm\)0.1 & 1.6\,\(\pm\)0.1 & 2.5\,\(\pm\)0.1 \\

ASER~\cite{shim2021ASER} & 24.8\,\(\pm\)2.2 & 32.9\,\(\pm\)1.2 & 41.7\,\(\pm\)1.7 & 12.2\,\(\pm\)0.4 & 15.7\,\(\pm\)0.8 & 19.5\,\(\pm\)0.5 & 2.1\,\(\pm\)0.2 & 3.4\,\(\pm\)0.5 & 7.1\,\(\pm\)0.7 \\

GDUMB~\cite{prabhu2020gdumb}  & 27.5\,\(\pm\)1.2 & 33.1\,\(\pm\)1.0 & 38.4\,\(\pm\)1.2 & 7.3\,\(\pm\)0.3 & 10.3\,\(\pm\)0.3 & 15.7\,\(\pm\)0.6 & 3.1\,\(\pm\)1.7 & 4.8\,\(\pm\)1.2 & 8.3\,\(\pm\)1.1 \\

DVC~\cite{DVC}  & 40.2\,\(\pm\)1.2 & 52.1\,\(\pm\)1.1 & 62.4\,\(\pm\)1.0 & 15.0\,\(\pm\)0.7 & 19.3\,\(\pm\)0.8 & 24.7\,\(\pm\)1.0 & 4.5\,\(\pm\)0.7 & 7.3\,\(\pm\)0.4 & 10.9\,\(\pm\)0.3 \\
SCR~\cite{mai2021SCR}  & 47.1\,\(\pm\)2.3 & 56.8\,\(\pm\)0.8 & 63.1\,\(\pm\)1.2 & 20.8\,\(\pm\)0.4 & 26.1\,\(\pm\)0.4 & 31.2\,\(\pm\)0.6 & 9.9\,\(\pm\)0.2 & 14.3\,\(\pm\)0.3 & 15.9\,\(\pm\)0.4 \\

PCR~\cite{lin2023pcr}           & 47.1\(\pm2.4\) & 56.8\(\pm0.8\) & 63.1\(\pm1.3\) & 20.8\(\pm0.4\) & 25.6\(\pm1.1\) & 31.2\(\pm0.6\) & 8.1\(\pm0.5\) & 12.3\(\pm0.7\) & 17.8\(\pm0.6\) \\

SSD~\cite{gu2024summarizing}        & 44.9\(\pm0.6\) & 57.8\(\pm0.6\) & 65.4\(\pm0.2\) & 21.2\(\pm0.4\) & 28.0\(\pm0.2\) & 31.9\(\pm0.5\) & 10.5\(\pm0.2\) & 15.7\(\pm0.1\) & 16.8\(\pm0.1\) \\
OCM~\cite{guo2022ocm}          & 57.1\(\pm1.9\) & 66.0\(\pm0.6\) & 69.7\(\pm1.1\) & 17.0\(\pm0.8\) & 23.7\(\pm0.5\) & 29.6\(\pm0.7\) & 10.0\(\pm0.3\) & 15.0\(\pm0.2\) &21.0\(\pm0.1\) \\
OnPro~\cite{wei2023onpro}          & 57.5\(\pm1.4\)  & 65.4\(\pm1.0\) & 70.0\(\pm0.5\) & 17.2\(\pm0.5\) & 22.3\(\pm0.7\) & 27.6\(\pm0.5\) &  7.8\(\pm0.4\) & 10.9\(\pm0.5\) & 16.2\(\pm0.3\) \\
EMI (Ours)           & \textbf{60.0}\(\pm1.4\) & \textbf{70.0}\(\pm0.8\) & \textbf{74.0}\(\pm0.6\) & \textbf{23.6}\(\pm0.8\)& \textbf{32.5}\(\pm0.5\) & \textbf{39.7}\(\pm0.4\) & \textbf{11.1}\(\pm0.5\)& \textbf{18.4}\(\pm 0.5\) & \textbf{24.0}\(\pm 0.3\) \\
\bottomrule

\end{tabular}
\end{table*}

\noindent
\textbf{Datasets.} We employ three image classification datasets:
\textit{CIFAR-10\cite{krizhevsky2009cifar}}, which contains 10 classes, with each class having 5000 training samples and 1000 test samples. We divide it into 5 tasks, each introducing two new classes.
\textit{CIFAR-100\cite{krizhevsky2009cifar}}, which has 100 classes, with each class having 500 training samples and 1000 test samples. We divide it into 10 tasks, each introducing ten new classes. 
\textit{Tiny-ImageNet\cite{le2015tinyimagenet}}, wihch has 200 classes. We split it into 100 tasks, each encompassing two non-overlapping classes, with 2,000 training samples and 200 testing samples per task.\\

\begin{figure*}[!htbp]
  
  \subfigure[CIFAR-10 ($M=200$)]{\includegraphics[width=0.245\textwidth]{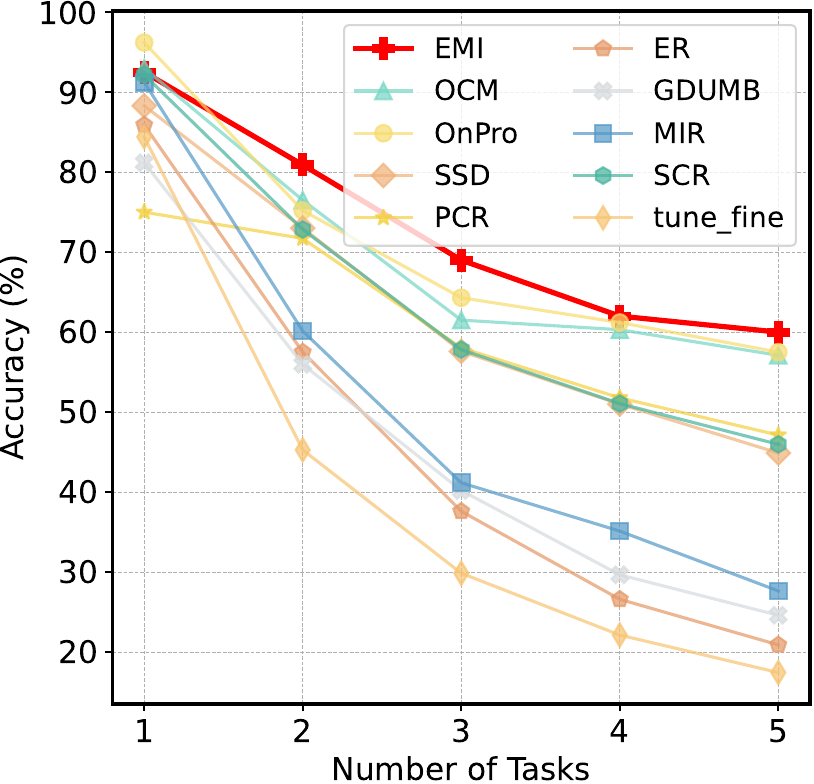}}
 \hfill 	
  \subfigure[CIFAR-10 ($M=500$)]{\includegraphics[width=0.245\textwidth]{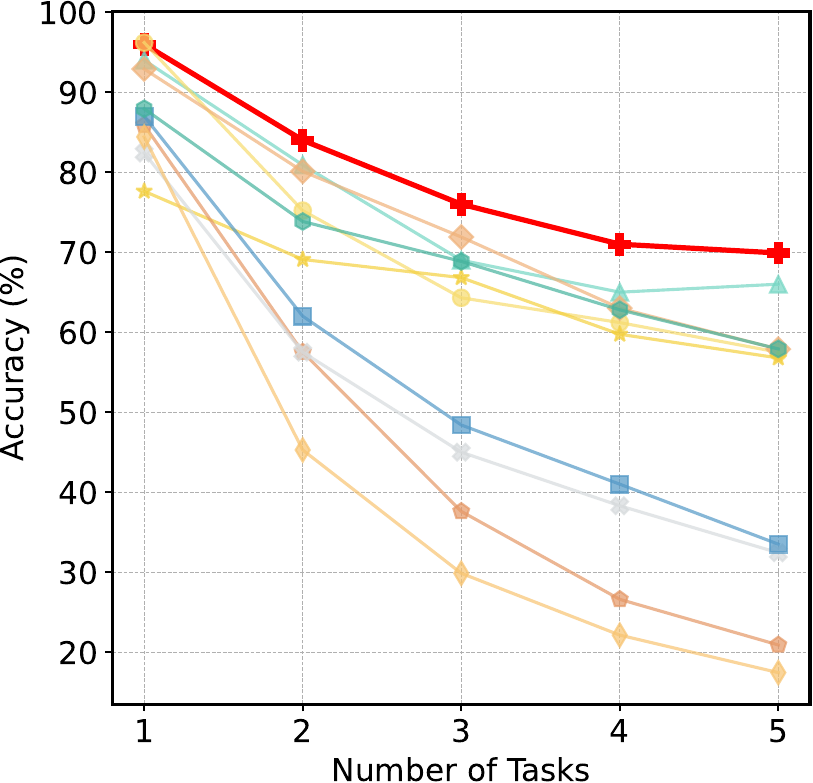}}
  \hfill
  \subfigure[CIFAR-100 ($M=500$)]{\includegraphics[width=0.245\textwidth]{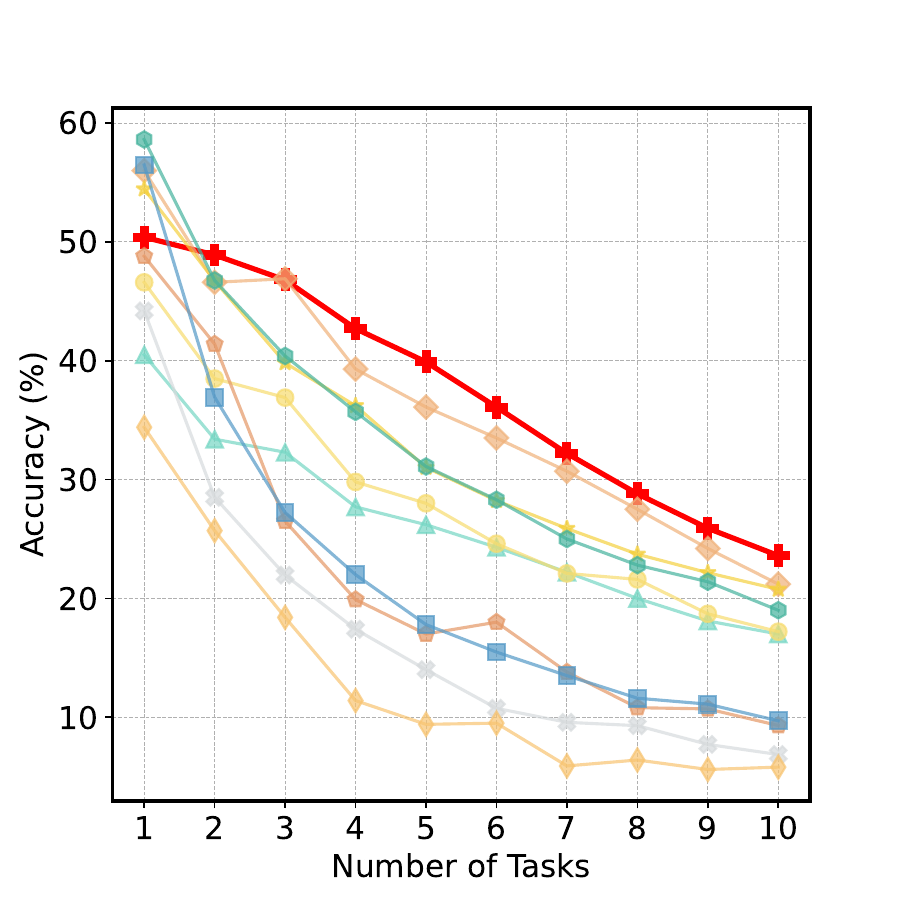}}
 \hfill 	
  \subfigure[CIFAR-100 ($M=1000$)]{\includegraphics[width=0.245\textwidth]{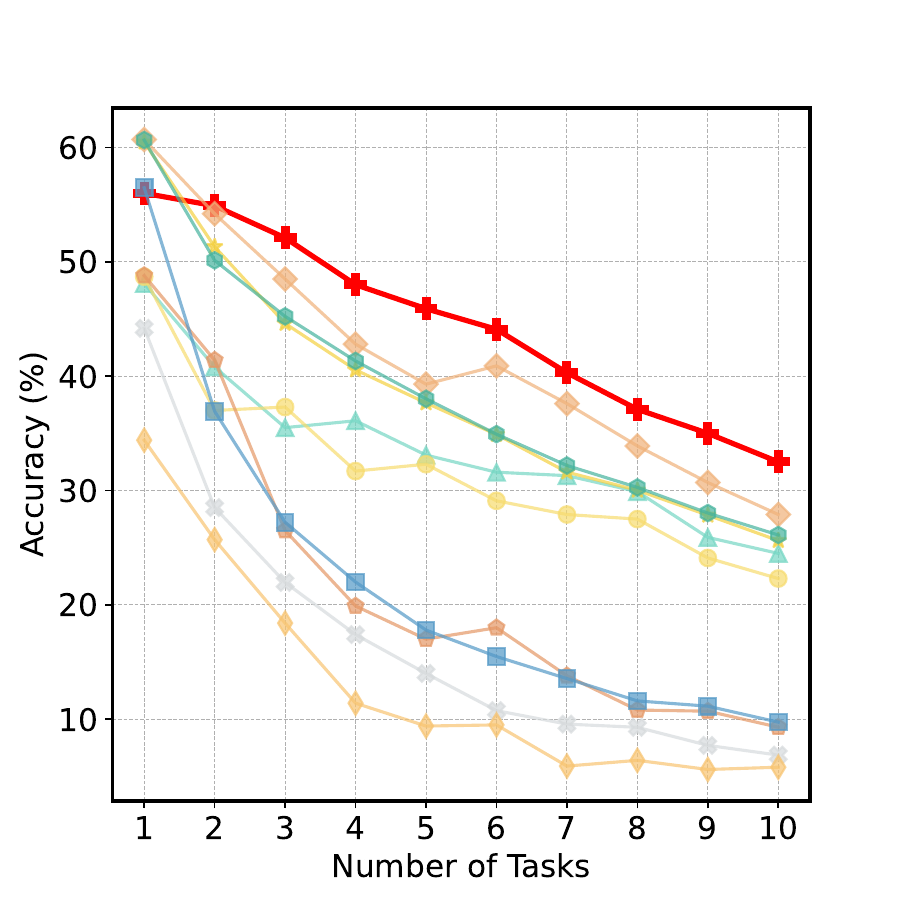}}
\caption{Incremental accuracy on tasks observed so far in the test set of CIFAR-10 and CIFAR-100 with different buffer sizes.}
\label{fig:IncreAC}
\end{figure*}

\noindent
\textbf{Methods to be compared.} We compare our method with some state-of-the-art OCIL methods, including AGEM \cite{chaudhry2018agem}, ER \cite{chaudhry2019ER}, MIR \cite{aljundi2019mir}, GSS \cite{aljundi2019gss}, ASER \cite{shim2021ASER}, DVC \cite{DVC}, GDUMB \cite{prabhu2020gdumb}, 
SCR \cite{mai2021SCR}, PCR \cite{lin2023pcr}, SSD \cite{gu2024summarizing}, OCM \cite{guo2022ocm}, OnPro \cite{wei2023onpro}. OCM is an MI-based method which is introduced in Eq .\eqref{eq:ocm}, and OnPro further develops the computation of online prototypes based on the partial MI relationships in OCM.
Moreover, we provide the experimental results of fine-tune and offline method. 
The fine-tune method trains the model sequentially as data arrives, without incorporating any strategies to prevent forgetting. 
The offline method trains the model through 30 epochs on the dataset in all datasets, utilizing i.i.d. sampled mini-batches, which defines the maximum performance achievable by such model.\\

\noindent
\textbf{Evaluation metrics.} We adopt Average Accuracy and Average Forgetting as metrics to measure our OCIL framework's performance. 
Average Accuracy reflects the model's accuracy over all observed tasks, computed as:
\begin{equation}
    \text{Average Accuracy} = \frac{1}{T} \sum_{j=1}^{T} a_{T,j},
\end{equation}
where \( a_{T,j} \) denotes the accuracy on task \( j \) subsequent to the model being trained up to task \( T \). Average Forgetting indicates the model's retention loss per task after it is trained on the final task, defined as:
\begin{equation}
    \text{Average Forgetting} = \frac{1}{T-1} \sum_{j=1}^{T-1} d_{T,j},
\end{equation}
where \( d_{T,j} \) quantifies the maximum accuracy decrement on task \( j \), calculated as:
\begin{equation}
    d_{i,j} = \max_{k \in \{1,\ldots,i-1\}} a_{k,j} - a_{i,j}.
\end{equation}
We first learn all tasks from the data stream for each dataset, and then test the final model using the test data of all tasks. We report the average accuracy of 10 random runs. \\

\noindent
\textbf{Implementation Details.} In training sequential tasks, following \cite{aljundi2019gss, chaudhry2018agem}, we employ a reduced ResNet-18 \cite{resnetHe} as the shared encoder, which is optimized by Adam, with an initial learning rate set at $10^{-3}$ for all datasets and a weight decay of $10^{-4}$.
The batch size (size of $\mathcal{X}$) is set to 10, as suggested in \cite{shim2021ASER}, and the replay batch size (size of $\mathcal{X}^b$) is set to 64 following \cite{guo2023dealing}.  
Moreover, the temperature $\tau$ is set to 0.07, and we experimentally determine that sampling $N_\text{p}=6$ samples per prototype yields the best performance. As for the similarity threshold $\mu$ in Eq .\eqref{eq:compute_similarity} for calculating $\mathcal{S}$, we adaptively computed as follows:
\begin{align}
\label{equ}
\mu = s_\text{max} - \alpha(s_\text{max} - s_\text{mean}),
\end{align}
where $s_\text{max}$ and $s_\text{mean}$ represent the maximum and mean values of pairwise similarities among intra-class samples, respectively, and $\alpha$ is a predefined hyperparameter. According to our experimental findings, we set $\alpha = 0.1$ for $\mathcal{S}$ and $\alpha_b = 0.2$ for $\mathcal{S}^b$.

For fairness comparisons, all baseline models are also implemented using the reduced ResNet-18 as the backbone, with identical batch and replay batch sizes. 
We reproduce all baseline models in a consistent environment using their source code and default settings.

\subsection{Comparisons with the State-of-the-arts}

\noindent
\textbf{Comparisons on average accuracy.} 
Tab. \ref{tab:ac} presents the average accuracy results of our EMI and all baseline methods with different buffer sizes $M$ on 3 benchmark datasets.
Across all datasets and buffer size settings, our EMI consistently outperforms all baselines.
On CIFAR-10, our method surpasses the second-best methods (OnPro and OCM) by 2.5\%, 4.6\%, and 4.0\% for buffer sizes of 0.2k, 0.5k, and 1k, respectively.
For CIFAR-100, with buffer sizes of 0.5k, 1k, and 2k, our method exceeds the second-best method (SSD) by 2.4\%, 4.5\%, and 7.8\%, respectively.
On Tiny-ImageNet, for buffer sizes of 1k, 2k, and 4k, our method outperforms the second-best methods (SSD and OCM) by 0.6\%, 2.7\%, and 3\%, respectively.
Several key observations can be summarized from the table: 
(1) EMI significantly outperforms other MI-based methods, such as OCM and OnPro. For example, on CIFAR-100 with  a buffer size of 2k, EMI achieves an accuracy that is 11.1\% and 12.1\% higher than OCM and OnPro, respectively. 
(2) EMI demonstrates considerable improvements as the buffer size is increased. For instance, on the Tiny-ImageNet dataset with a buffer size of 0.5k, our approach achieves an accuracy of 11.1\%, close to OCM, which records a 10.0\% accuracy. Significantly, as the buffer size is expanded to 4k, our method exhibits a remarkable increase of 12.9\%. In comparison, OCM, despite starting from a lower baseline accuracy, shows an increase of only 11.0\%. 
(3) Many continual learning methods experience a drastic performance drop as the number of tasks increases. For instance, OnPro achieves only a 16.8\% accuracy on Tiny-ImageNet with 100 tasks, even with a 4k buffer size. In contrast, EMI demonstrates exceptional adaptability to increasing task numbers. Across all buffer sizes on Tiny-ImageNet, it maintains strong performance, notably achieving 24.0\% with a 4k buffer—nearly matching the offline performance benchmark of 25.9\%.\\

\begin{table*}[t]
\centering
\caption{Average forgetting rate (lower is better) on three benckmark datasets. All results are the average and standard deviation of 10 runs.}
\label{tab:af}
\begin{tabular}{l|ccc|ccc|ccc}
\toprule
\multirow{2}{*}{\textbf{Method}} & \multicolumn{3}{c|}{\textbf{CIFAR-10} (5 tasks)} & \multicolumn{3}{c|}{\textbf{CIFAR-100} (10 tasks)} & \multicolumn{3}{c}{\textbf{Tiny-ImageNet} (100 tasks)} \\
                        & \(M=0.2k\) & \(M=0.5k\) & \(M=1k\) & \(M=0.5k\) & \(M=1k\) & \(M=2k\) & \(M=1k\) & \(M=2k\) & \(M=4k\) \\ \midrule

AGEM~\cite{chaudhry2018agem}                  &67.8\,\(\pm\)3.0 & 69.2\,\(\pm\)5.1 & 69.6\,\(\pm\)0.6 & 44.6\,\(\pm\)0.7 & 44.2\,\(\pm\)0.5 & 44.8\,\(\pm\)1.4 &  74.3\,\(\pm\)0.3 & 73.7\,\(\pm\)0.3 & 74.0\,\(\pm\)0.2\\

ER~\cite{chaudhry2019ER} & 67.1\,\(\pm\)0.5 &62.7\,\(\pm\)0.7 & 52.8\,\(\pm\)1.8 & 45.4\,\(\pm\)0.4 & 77.5\,\(\pm\)1.3 & 78.3\,\(\pm\)1.2 & 78.2\,\(\pm\)0.1 & 76.1\,\(\pm\)0.2 & 74.1\,\(\pm\)0.3 \\
MIR~\cite{aljundi2019mir} & 59.5\,\(\pm\)0.8 & 53.2\,\(\pm\)0.6 & 39.3\,\(\pm\)1.0 & 51.6\,\(\pm\)0.1 & 49.4\,\(\pm\)0.4 & 45.8\,\(\pm\)0.2 & 77.9\,\(\pm\)0.2 & 73.9\,\(\pm\)0.4 & 67.4\,\(\pm\)0.6 \\
GSS~\cite{aljundi2019gss} & 68.9\,\(\pm\)0.7 & 59.1\,\(\pm\)1.2 & 47.7\,\(\pm\)1.0 & 49.9\,\(\pm\)0.2 & 45.2\,\(\pm\)0.6 & 40.1\,\(\pm\)0.6 & 76.3\,\(\pm\)0.2 & 74.2\,\(\pm\)0.2 & 70.7\,\(\pm\)0.3 \\
ASER~\cite{shim2021ASER} & 64.6\,\(\pm\)0.7 & 56.2\,\(\pm\)1.2 & 45.8\,\(\pm\)1.2 & 49.2\,\(\pm\)0.2 & 44.2\,\(\pm\)0.3 & 38.3\,\(\pm\)0.2 & 76.8\,\(\pm\)0.3 & 73.0\,\(\pm\)0.2 & 64.1\,\(\pm\)0.3 \\

GDUMB~\cite{prabhu2020gdumb}  & 22.3\,\(\pm\)0.7 & 19.9\,\(\pm\)0.6 & 20.5\,\(\pm\)0.5 & 23.2\,\(\pm\)0.4 & 22.2\,\(\pm\)0.3 & 19.4\,\(\pm\)0.2 & 24.9\,\(\pm\)1.4 & 23.1\,\(\pm\)0.5 & 23.9\,\(\pm\)0.5 \\

DVC~\cite{DVC}  & 28.9\,\(\pm\)1.9 & 18.4\,\(\pm\)2.0 & 15.4\,\(\pm\)1.5 & 27.9\,\(\pm\)0.7 & 26.3\,\(\pm\)1.8 & 21.9\,\(\pm\)2.0 & 26.5\,\(\pm\)0.7 & 23.3\,\(\pm\)0.9 & 19.9\,\(\pm\)0.2 \\
SCR~\cite{mai2021SCR}   & 32.1\,\(\pm\)3.0 & 19.9\,\(\pm\)1.6 & 17.3\,\(\pm\)0.4 & 23.0\,\(\pm\)0.2 & 17.1\,\(\pm\)0.1 & 10.1\,\(\pm\)0.1 & 22.1\,\(\pm\)0.2 & 18.1\,\(\pm\)0.2 & 14.3\,\(\pm\)0.1 \\

PCR~\cite{lin2023pcr}         & 28.3\(\pm1.0\) & 18.8\(\pm0.6\) & 13.1\(\pm0.6\) & 30.6\(\pm0.3\) & 23.8\(\pm0.4\) & 17.3\(\pm0.3\) & 26.7\(\pm1.0\) & 23.9\(\pm0.5\) & 22.1\(\pm0.5\) \\
SSD~\cite{gu2024summarizing}         & 41.8\(\pm0.6\) & 28.9\(\pm0.6\) & 18.2\(\pm0.5\) & 22.3\(\pm0.4\) & 16.2\(\pm0.1\) & 10.2\(\pm0.2\) & 20.9\(\pm0.3\) & 16.8\(\pm0.2\) & 13.0\(\pm0.1\) \\
OCM~\cite{guo2022ocm}            & 22.9\(\pm1.0\) & 14.5\(\pm0.7\) & 12.1\(\pm0.6\) & \textbf{13.5}\(\pm0.4\) & \textbf{8.6}\(\pm0.3\) & \textbf{4.4}\(\pm0.2\) & 23.6\(\pm0.3\) &19.2\(\pm0.2\) & 15.8\(\pm0.2\) \\
OnPro~\cite{wei2023onpro}           & 23.2\(\pm0.9\)  & 16.4\(\pm0.7\) & 12.7\(\pm0.5\) & 17.1\(\pm0.2\) & 9.6\(\pm0.2\) & 6.1\(\pm0.2\) & 21.0\(\pm0.3\) & 19.2\(\pm0.2\) & 17.0\(\pm0.3\) \\
EMI (Ours)            & \textbf{21.1}\(\pm0.7\) & \textbf{13.1}\(\pm0.5\) & \textbf{9.0}\(\pm0.3\) & 19.3\(\pm0.3\) & 13.6\(\pm0.3\) &8.9\(\pm0.2\) & \textbf{20.1}\(\pm0.2\) & \textbf{16.3}\(\pm0.3\) & \textbf{12.4}\(\pm0.2\) \\
\bottomrule
\end{tabular}
\end{table*}

\noindent
\textbf{Comparisons on average forgetting.} 
We also report the average forgetting results of our EMI and all baselines in Tab. \ref{tab:af}.
For CIFAR-10 and Tiny-ImageNet datasets, our EMI has the lowest average forgetting rates compared to all replay-based baseline models. 
For the CIFAR-100 dataset, our results are higher than OCM and OnPro but lower than all other methods. The reason for this is that our adoption of the DualNet architecture and its optimization for both fast and slow features, which not only deepens the model's understanding of inter-class general knowledge but also enhances its grasp of intra-class specific knowledge. Consequently, due to the model's engagement with a greater volume of learned knowledge compared to other methods, it inherently exhibits a higher rate of forgetting.
Despite this, our method still retains more knowledge than OCM and OnPro.

In terms of average accuracy, our method performs significantly better than OCM and OnPro. 
For CIFAR-100, our method exceeds OCM by 6.6\%, 8.8\%, and 9.9\% for buffer sizes of 0.5k, 1k, and 2k, respectively. Compared to OnPro, our method shows improvements of 6.4\%, 9.3\%, and 12.1\% for the same buffer sizes. It is important to note that when the maximum achievable accuracy for a task is inherently low, the observed forgetting rate appears minimal, even if the model completely discards the learned content. This phenomenon occurs because the baseline performance on such tasks is already low, making the impact of forgetting relatively insignificant. 
Moreover, since OCM and OnPro are both based on MI, and our method also employs a MI-based approach, the forgetting rates of these three methods are significantly lower than other baselines. This demonstrates that MI methods can effectively extract the overall features within constructed relationships, thereby reducing knowledge forgetting.\\

\noindent
\textbf{Comparisons of incremental training process.}
Fig. \ref{fig:IncreAC} illustrates the incremental learning performance of various methods on the CIFAR-10 and CIFAR-100 datasets under different memory sizes (0.2k, 0.5k, and 1k).
From the variations in the curves across the four graphs, two key observations can be made:
(1) The EMI method consistently maintains relatively high accuracy across all tasks and effectively mitigates the problem of forgetting. In contrast, the performance of most baseline methods deteriorates rapidly with the addition of new classes.
(2) The advantage of the EMI method is particularly pronounced with larger memory sizes. Specifically, for CIFAR-10 with M=500 and CIFAR-100 with M=1000, EMI significantly outperforms the second-best method at almost every incremental stage.
These findings indicate that EMI demonstrates superior robustness and adaptability in managing memory knowledge and progressively increasing tasks. 
{Additionally, we observe that the accuracy of the EMI method in the first task of CIFAR-100 is not the best. This is because the EMI method focuses on memory and knowledge decomposition across tasks, which means the advantage of MI is not evident in the first task (the accuracy of other MI-based methods is also low in the first task). However, in the subsequent tasks, the accuracy of EMI is consistently the highest.}

\begin{figure*}[t]
  \subfigure[ER~\cite{chaudhry2019ER}]{\includegraphics[width=0.49\textwidth]{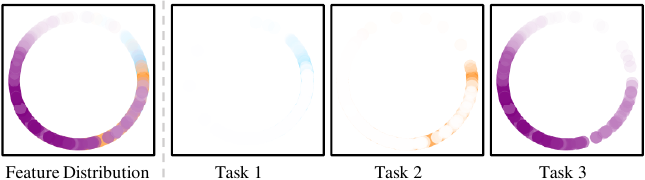}}
 \hfill 	
  \subfigure[OCM~\cite{guo2022ocm}]
  {\includegraphics[width=0.49\textwidth]{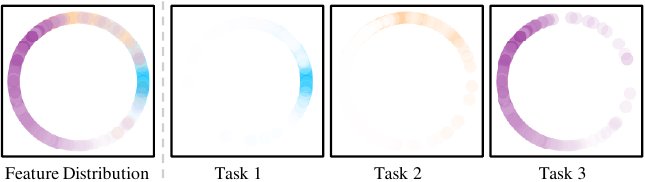}}
   \hfill 	
  \subfigure[OnPro~\cite{wei2023onpro}]
  {\includegraphics[width=0.49\textwidth]{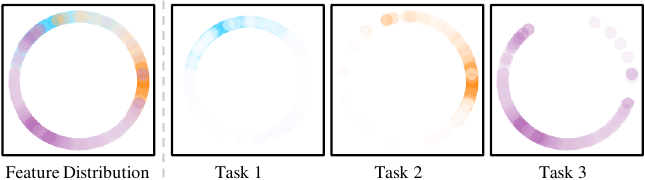}}
   \hfill 	
  \subfigure[EMI (Ours)]
  {\includegraphics[width=0.49\textwidth]{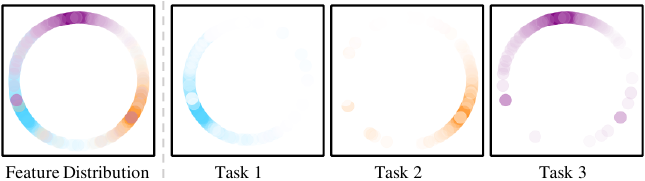}}
\caption{We plot feature distributions with Gaussian kernel density estimation (KDE) in $\mathbb{R}^2$ and visualizes the distributions on a unit circle. Three rightmost plots visualize feature distributions of selected specific tasks. The representation from EMI is evenly distributed, enabling it to learn a more uniform feature representation.}
\label{fig:distribution_angles}
\end{figure*}

\begin{figure*}[t]
  \subfigure[ER~\cite{chaudhry2019ER}]{\includegraphics[width=0.245\textwidth]{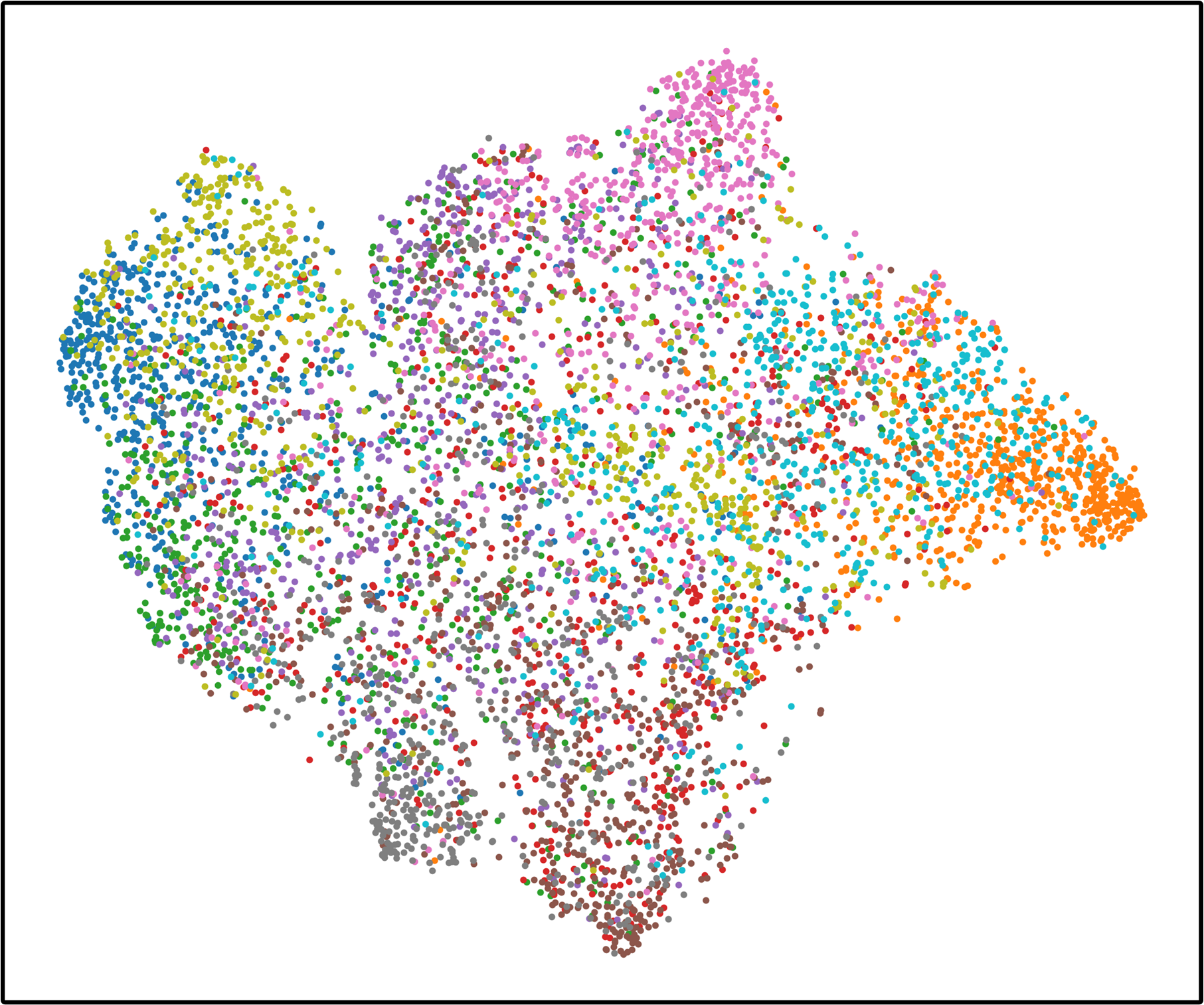}}
 \hfill 	
  \subfigure[OCM~\cite{guo2022ocm}]{\includegraphics[width=0.245\textwidth]{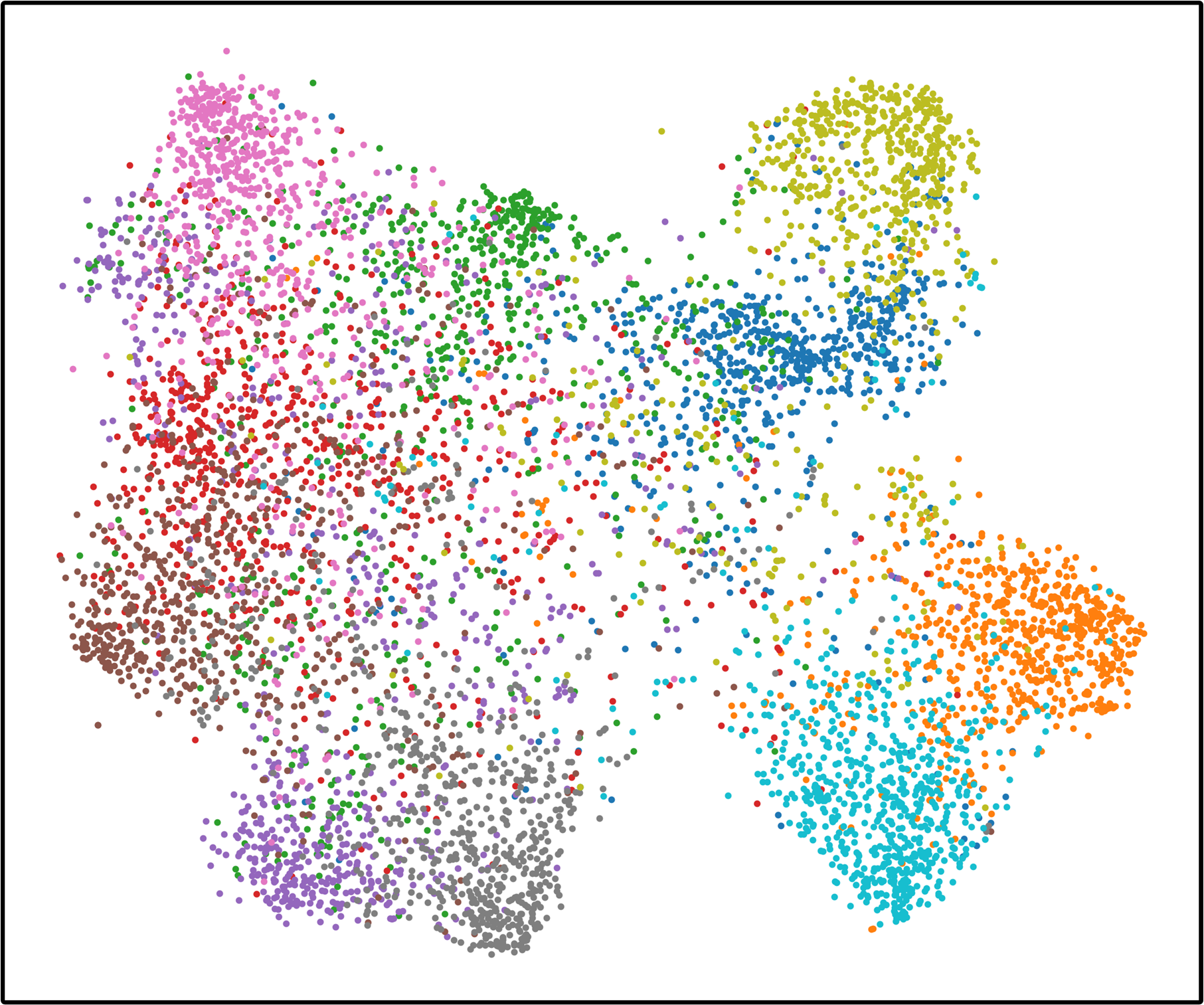}}
 \hfill 	
  \subfigure[OnPro~\cite{wei2023onpro}]{\includegraphics[width=0.245\textwidth]{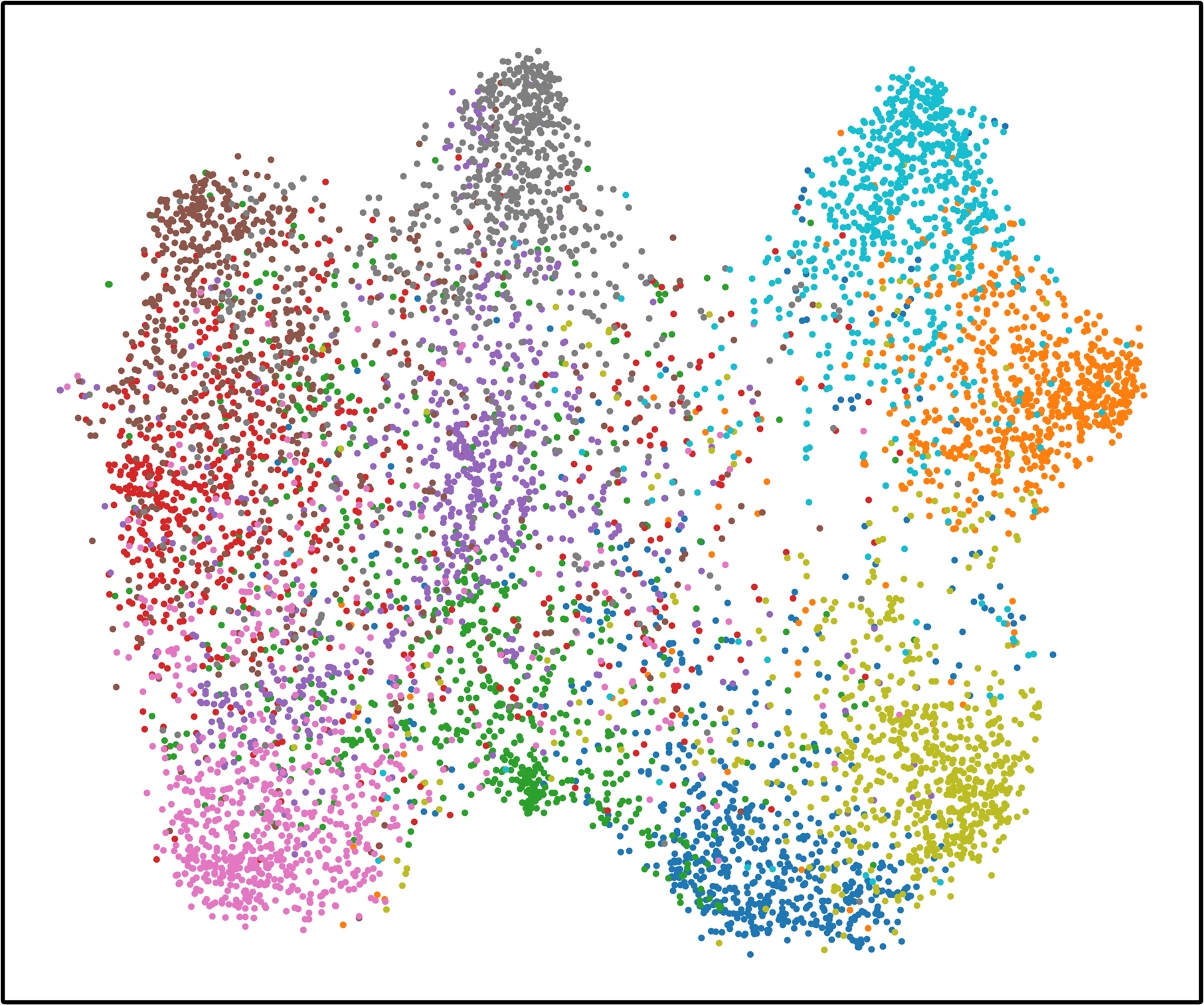}}
   \hfill 	
  \subfigure[EMI (Ours)]{\includegraphics[width=0.245\textwidth]{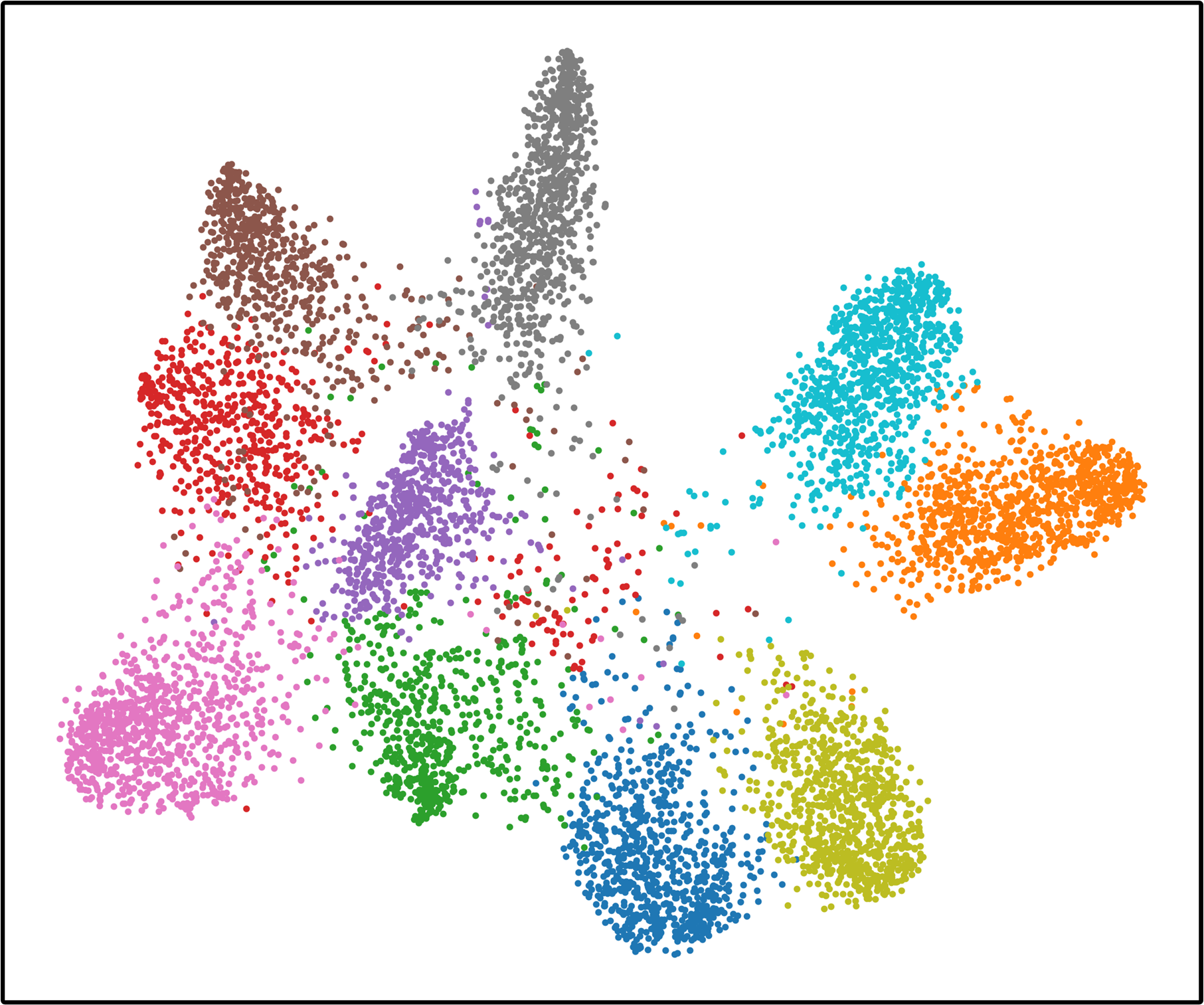}}
\caption{t-SNE visualizations of features learned on the test set of CIFAR-10. Compared to other methods, the intra-class distribution of EMI is more compact, and the boundaries between classes are clearer.}
\label{fig:tsne}
\end{figure*}

\subsection{Feature Distribution Visualization} 
\noindent
\textbf{Diversity visualization}.
In OCIL, the network cannot simultaneously view all data, and each data point is only seen once. This makes it difficult to learn a uniform feature distribution on the hyperspherical surface, leading to confusion in the network's learned knowledge. 
To intuitively demonstrate this issue, we plot feature distributions using kernel density estimation (KDE) in $\mathbb{R}^2$ and visualize the distributions on angles in the training set of CIFAR-10 (M=1k) after the model is incrementally trained, as shown in Fig. \ref{fig:distribution_angles}. 
We observe that the feature distribution of ER is highly uneven. The features from the most recent task (task 3) occupy approximately three-quarters of the space, with a broader angle distribution and stronger responses compared to earlier tasks (task 1 and task 2). However, methods based on MI, such as OCM and OnPro, alleviate this phenomenon to some extent, though task 3 still occupies half of the feature space.
In contrast, in our proposed EMI, the distribution of features from the three tasks is much more uniform, with clear boundaries between the distributions of different tasks. 
This is because DMI leverages inter-class samples to construct MI, enabling the network to learn richer features, particularly edge and variant features. This results in a more reasonable distribution of different data and tasks, leading to a more balanced overall data distribution. \\

\noindent
\textbf{Representativeness and separability visualization}.
Fig. \ref{fig:tsne} displays t-SNE visualizations of features learned on the test set of CIFAR-10 ($M=1k$) using different methods: ER, OCM, OnPro, and EMI. 
For ER, the feature distribution is highly disordered, characterized by overlapping feature distributions between classes and the inability of intra-class sample features to form compact clusters. 
In contrast, MI-based methods such as OCM and OnPro demonstrate better performance. Their intra-class sample distributions generally occupy distinct regions separate from other classes. However, their distributions are still not compact enough, and the class boundaries remain indistinct.
Our proposed EMI method addresses these issues effectively. 
By utilizing RMI, we align intra-class samples with the representative features of each class, resulting in a more compact intra-class distribution. Furthermore, we employ SMI to achieve clear separation between inter-class distributions, thereby ensuring more distinct class boundaries.

\begin{table}[t]
\centering
\caption{Ablation studies on CIFAR-100 (M = 1k). 
"DualNet" means using DualNet as backbone and the loss is  $\mathcal{L}_\text{ocm}$.}
\label{tab:abl-1}
\scalebox{1}{%
\begin{tabular}{cccccc}
\toprule
 DualNet& $\mathcal{L}_\text{dmi}$ & $\mathcal{L}_\text{rmi}$ & $\mathcal{L}_\text{smi}$ & \(\text{Accuracy}\)\\ \midrule
  &  & & & 23.7   \\
 \checkmark&  &   & & 26.8  \\
\checkmark  &\checkmark  & & & 30.1 \\
\checkmark&  &\checkmark &  & 29.1   \\
\checkmark& \checkmark&\checkmark &  & 32.0   \\
\checkmark& \checkmark&  &\checkmark  & 31.1   \\
\checkmark& \checkmark&\checkmark &\checkmark  &32.5   \\
\bottomrule

\end{tabular}%
}
\end{table}
\subsection{Ablation Study}
We conduct ablation experiments to analyze the contribution of various components and choices made in EMI with 1k ($M$=1k). The results are given in Tab. \ref{tab:abl-1}\\

\noindent
\textbf{Ablation study for each compent}.
In our experiments, we use a simplified ResNet as the backbone network and applied the loss function $\mathcal{L}_\text{ocm}$. "DualNet," mentioned in the second row of Tab. \ref{tab:abl-1}, refers to a backbone network composed of a simplified ResNet and a parallel four-layer CNN. In this setup, $\mathcal{L}_\text{ocm}$ only optimizes fast features, without any loss function optimizing the slow features, resulting in a 2.9\% increase in accuracy. This improvement is attributed to the addition of a Slow Net for knowledge transform. However, utilizing only fast features does not fully leverage the potential of the DualNet architecture.
The results in the third row indicate that when DMI is used to optimize slow features, accuracy increases to 30.1\%, an improvement of 6.4\% over the baseline. This demonstrates that $\mathcal{L}_\text{dmi}$ significantly enhances the depth and breadth of the learning process, contributing to the acquisition of a more diverse feature set and underscoring the necessity of optimizing slow features.
We also presents the results of an ablation study on RMI in the fourth row. To evaluate the effectiveness of RMI, we excluded the DMI component, meaning we did not specifically optimize for slow features. The results show a 5.4\% improvement over baseline, validating the effectiveness of RMI. However, this is 1.0\% lower than the result with the DMI component included (third row), again highlighting the importance of optimizing slow features. Furthermore, when RMI and DMI are combined (fifth row), performance reaches 32.0\%, an 8.3\% improvement over our baseline. This is because the more general features learned by DMI are effectively utilized by RMI, creating a synergistic effect where the combined impact exceeds the sum of their individual contributions.
Finally, by integrating all the aforementioned methods, we considered diversity, representativeness, and separability in constructing the MI. As shown in the last row, we achieved the best result, improving by 8.8\% over our baseline, reaching a score of 32.5\%.\\

\begin{table}[t]
\centering
\caption{Parameter analysis on the CIFAR-100 is performed for various $\alpha$ and $\alpha_b$ values in DMI. Here, $\alpha$ and $\alpha_b$ specifically are the parameters for calculation $\mathcal{S}$ and $\mathcal{S}^b$, respectively. Given the different distributions of $\mathcal{X}$ and $\mathcal{X}^b$, these parameters are considered separately.}
\label{tab:para_DMI}
\begin{tabular}{ccccc}
\toprule
\textbf{} & $\alpha_b = 0.0$ & $\alpha_b = 0.2$ & $\alpha_b = 0.4$ & $\alpha_b = 0.6$ \\
\midrule
$\alpha = 0.0$ & 31.7 & 32.2 & 32.0& 31.8 \\
$\alpha = 0.1$ & 32.2  & \textbf{32.5} & 32.3 & 32.0 \\
$\alpha = 0.2$ & 32.0  & 32.3  & 32.1  & 31.7 \\
$\alpha = 0.3$ & 32.0  & 32.0  & 31.8  & 31.2 \\
\bottomrule
\end{tabular}
\end{table}

\begin{table}[t]
\centering
\caption{Parameter analysis on the CIFAR-100 dataset is performed for various $N_\text{p}$ values in RMI. This table displays the results in terms of Accuracy.}
\label{tab:para_RMI}
\begin{tabular}{cccccc}
\toprule
$N_\text{p}$  & 1  & 3  & 6  & 9 & 12\\
\midrule
Accuracy  & 31.5  & 32.0  & 32.5 & 32.3 & 32.1 \\
\bottomrule
\end{tabular}
\end{table}

\begin{table}[t]
\centering
\caption{The efficiency analysis experiment is conducted on CIFAR-10 and compared with other MI-based methods. We have recorded the GPU memory usage and the time consumed for each batch during training.}
\label{tab:effic_ana}
\begin{tabular}{cccc}
\toprule
  & OCM~\cite{guo2022ocm}  & OnPro~\cite{wei2023onpro}  & EMI (Ours)\\
\midrule
Training Time (ms)  & 61.6  & 92.9  & 84.1 \\
Memorey (GB)& 3.2  & 4.5  & 3.8\\
\bottomrule
\end{tabular}
\end{table}

\noindent
\textbf{Parameter analysis for DMI}.
Additionally, we conduct a parameter experiment for $\alpha$ in DMI, as shown in Tab. \ref{tab:para_DMI}. This experiment was performed on the CIFAR-100 dataset with $M$ = 1k. Notably, when both $\alpha$ and $\alpha_b$ are set to 0, DMI constructs MI only intra-class, resulting in lower performance compared to when $\alpha$ and $\alpha_b$ are nonzero. This indicates that DMI effectively guides the learning of richer features. The model achieves the highest accuracy when $\alpha = 0.1$ and $\alpha_b = 0.2$. However, further increases in $\alpha$ and $\alpha_b$ deteriorate the network's performance. This is attributed to the overly lenient similarity evaluation standards, which disrupt the balance between intra-class and inter-class samples. This imbalance leads to DMI with insufficient similarity, causing the network to learn incorrect knowledge and leading to a decline in performance.\\

\noindent
\textbf{Parameter analysis for RMI}.
We conduct an experiment to evaluate the parameter $N_\text{p}$ in the RMI module in Tab. \ref{tab:para_RMI}. For prototype calculation, we sampled $N_\text{p}$ instances from the buffer for each class in the data stream. The results highlight three key observations: (1) Even with a single sample ($N_\text{p}$=1), there is a notable performance improvement of 0.4 points compared to not using RMI (sixth row in Tab. \ref{tab:abl-1}). (2) The network achieves optimal performance when $N_\text{p}$ is set to 6, showing a 1.0 percentage point improvement over $N_\text{p}=1$. (3) Increasing $N_\text{p}$ beyond 6 does not yield better results; in fact, performance degradation is observed. We hypothesize that larger $N_\text{p}$ values introduce excessive noise into the prototypes, thereby affecting the network's performance.\\

\noindent
\textbf{Efficiency analysis}.
We conduct an efficiency analysis experiment. As shown in the Tab. \ref{tab:effic_ana}, the GPU memory usage and time consumed per batch for training were recorded for three MI-based methods: OCM, OnPro, and EMI.
Comparing the memory efficiency, OCM shows the lowest memory usage at 3.2 GB, followed by EMI at 3.8 GB, and OnPro at the highest with 4.5 GB. Regarding time efficiency, OCM is the fastest with 61.6 ms per batch, EMI follows with 84.1 seconds, and OnPro is the slowest at 92.9 seconds.
Although EMI's efficiency is moderate (worse than OCM but better than OnPro), it offers significantly higher accuracy than both methods, making it a valuable approach.

\section{Conclusion and Future Work}

In this paper, we propose a novel framework EMI, which is designed to decouple knowledge in OCIL for the sake of mitigating forgetting and improve novel plasticity.
EMI builds MI from the perspectives of diversity, representativeness, and separability. 
Specifically, we first employ inter-class samples to construct DMI and maximize it to achieve diversified knowledge. 
Then, we utilize prototypes to represent the representative features of the data stream. By constructing RMI between samples and prototypes as well as SMI between prototypes, we achieve representative knowledge and separability knowledge. These two types of knowledge ensure the consistency of intra-class knowledge and the differentiation of inter-class knowledge.
In the feature space, this results in a more compact distribution of intra-class features and clearer decision boundaries between inter-class features, thereby enhancing the stability and discriminative power of the network.
The experiments on several benchmark datasets show that the proposed EMI achieve new SOTA results in OCIL.

Despite the significant effectiveness of the EMI framework, the need to simultaneously optimize fast and slow features may impose certain limitations on processing speed. This limitation could affect the framework's applicability in scenarios that requires real-time training. In the future, we plan to improve the processing efficiency of our EMI framework to further enhance its performance in real-world applications.

\bibliographystyle{IEEEtran}
\bibliography{refs}

\end{document}